\documentclass{article}
\usepackage[preprint]{tmlr}

\usepackage{hyperref}
\usepackage{url}
\usepackage{algorithm}
\usepackage{algpseudocode}

\usepackage{graphicx} % Required for inserting images
\usepackage{comment}
\usepackage{amsfonts,bm,amsmath,amsthm,amssymb}
\usepackage{url}
\usepackage{xcolor}
\usepackage{subcaption}
\theoremstyle{definition}

\theoremstyle{definition}

\newtheorem*{exercise*}{Exercise}
\theoremstyle{remark}

\usepackage{soul}
\usepackage[english]{babel}
\usepackage{pgfplots}
\usepackage{tikz}
\DeclareUnicodeCharacter{0229}{e}

\title{
Mixing Artificial and Natural Intelligence:\\ 
{\large From Statistical Mechanics to AI and Back to Turbulence}}
\author{Michael (Misha) Chertkov\\
Program in Applied Mathematics, University of Arizona, Tucson\\
E-mail address for corresponding author: chertkov@arizona.edu
}
\date{\today}

\begin{document}

\maketitle

\begin{abstract} 
The paper reflects on the future role of AI in scientific research, with a special focus on turbulence studies, and examines the evolution of AI, particularly through Diffusion Models rooted in non-equilibrium statistical mechanics. It underscores the significant impact of AI on advancing reduced, Lagrangian models of turbulence through innovative use of Deep Neural Networks. Additionally, the paper reviews various other AI applications in turbulence research and outlines potential challenges and opportunities in the concurrent advancement of AI and statistical hydrodynamics. This discussion sets the stage for a future where AI and turbulence research are intricately intertwined, leading to more profound insights and advancements in both fields.
\end{abstract}

I wrote this paper to articulate my personal view on the relationship between Artificial Intelligence (AI) and quantitative scientific disciplines, with a strong focus on turbulence within the realm of fluid dynamics. 

I begin in Section \ref{sec:intro} by defining AI, highlighting key components such as Automatic Differentiation, Deep Learning, Reinforcement Learning, and Generative Models, and discussing their progressive interdependence \footnote{  In this manuscript, AI is used as a collective term to encompass various techniques, including but not limited to Machine Learning (ML). ML is a subset of AI focused on algorithms that enable computers to learn from and make predictions based on data.} This preliminary discussion also helps to set the stage for methodologies of how we use AI (as System 1 or System 2) and explain how AI's capacity to navigate the curse of dimensionality has sparked significant interest in scientific communities, enabling novel approaches to longstanding computational challenges.

In Section \ref{sec:my-sciences} I describe ``my sciences" to give the reader a glimpse on  why and how I see the disciplines as distinct as AI, theoretical engineering  and turbulence connected. 

Section \ref{sec:AIforSc} examines the evolving role of AI in scientific discovery. It discusses how "Physics-Agnostic AI" can uncover physical principles without explicit programming, the development and significance of "Physics-Informed AI" where physical laws are integrated into AI models, and the potential of AI in the future to autonomously discover new laws of physics. This Section reflects on AI's current contributions to understanding physical phenomena and its promising trajectory towards deepening and expanding our scientific knowledge base.

Section \ref{sec:diff} is devoted to our recent exploration of the diffusion models -- a class of generative AI models that is based on non-equilibrium statistical mechanics, which have remarkable abilities to generate high-fidelity simulations from learned data distributions, and shares many common features, which are far from being explored, with turbulence. 

In Section \ref{sec:turb}, we pivot towards turbulence/physics, initiating with a primer on Lagrangian Turbulence, where we trace the trajectories of fluid particles. We advocate for the Lagrangian method as a bridge linking diffusion models with turbulence studies, providing deeper insights into the intricate, multi-scale dynamics of turbulence. This approach, beneficial in pre-AI turbulence research, led to the development of a Lagrangian closure model encapsulated by a tetrad that represents the dynamics of a fluid element.

Advancing to the AI era, we explore how, two decades later, AI tools have rejuvenated this field. Specifically, we delve in Section \ref{sec:VG} into the application of AI resulting in a ``Neural Turb ODE" model offering refined closure for modeling velocity gradients in turbulence. Section \ref{sec:Eulerian} then takes a detour to discuss and classify the landscape of AI models enhancing Eulerian Turbulence. The narrative culminates in Section \ref{sec:multi}, which discusses the adaptation of Smooth Particle Hydrodynamics (SPH) to the Neural Lagrangian Large Eddy Simulation (L-LES) framework, marking a significant stride in integrating AI with traditional turbulence modeling techniques. This integration not only honors the Lagrangian perspective's conservation laws but also paves the way for future AI-augmented turbulence models that can seamlessly blend with and enhance classical computational fluid dynamics methodologies.

\section{Instead of Introduction}
\label{sec:intro}

\subsection{Disclaimer}

The inception of these notes was under a tentative title, {\it "What is Artificial Intelligence and How it Depends On and Affects Sciences?"}, which, although it has since evolved, remains pertinent to mention. Initially intended primarily for personal use or within my research group, I ultimately realized that the broader scientific community might also find value in the insights and discussions presented herein.

\subsection{What is AI? Many Opinions:}
In today's discourse, the term "Artificial Intelligence" (AI) frequently dominates conversations. However, the interpretation and significance attached to this compact abbreviation vary greatly among individuals.

\subsection{The AI Landscape}
My use of the term is technical. The AI I discuss includes:
\begin{itemize}
%    \item Automatic Differentiation: This involves speeding up the basic elements of analog computation using a set of elementary operations and the chain rule. It enables accurate and efficient gradient evaluations for optimization and machine learning. (For a comprehensive review, consult \cite{baydin_automatic_2015}.)

    \item {  Automatic Differentiation: often abbreviated as AD, is a set of techniques to evaluate the derivative of a function specified by a computer program. Unlike numerical differentiation (which approximates derivatives using finite differences and can be prone to numerical errors) or symbolic differentiation (which manipulates complex mathematical expressions to find derivatives and can be computationally intensive), AD provides a way to calculate derivatives exactly and efficiently by systematically applying the chain rule of calculus. (For a comprehensive review, consult \cite{baydin_automatic_2015,lavin_simulation_2022}.)}
    
    \item Deep Learning: It encompasses the most efficient parametric representation of continuous-valued functions, as compositions of parameterized, nonlinear transformations, using Neural Networks with many layers. It enables the approximation of complex relationships within data through iterative optimization -- thus learning -- of these (many billions) of parameters.
    (Refer to the seminal work for foundational insights \cite{lecun_deep_2015}.)
    
    \item Reinforcement Learning: This is a data-driven approach to implementing optimal control under uncertainty. It is an adaptive approach in machine learning, which reinforces decisions based on the information, e.g. reward, received in the process of learning (exploration), to get better results (inference, also called exploitation. (For a comprehensive guide, refer to \cite{sutton_reinforcement_2018}.)
    
    \item Generative Models: This refers to AI models and algorithms that can generate new, previously unseen data that is either similar in statistical properties to the training data or complete (answer the questions posed in) input data. (For further reading on the subject, refer to \cite{OpenAI_ChatGPT}.)
\end{itemize}

The list is progressive in the sense that each next item depends on, or at least related to, the previous one.

\subsection{The Excitement Surrounding AI}

In my opinion, the excitement within the scientific community surrounding AI stems from its unique ability to tackle the challenges posed by the curse of dimensionality in numerous scientific domains heavily reliant on computations. AI has illuminated the notion that computational hardness, or the inability to efficiently solve problems within existing frameworks and models, is not a hindrance. Instead, it serves as an invitation to circumvent these limitations and craft remarkable tools. One major lesson we have learned from AI is that when faced with unsolvable problems (theoretically and/or practically), we can change the formulations and borrow approaches and methods from other disciplines, notably applied mathematics and statistical physics. A prime example of this ingenuity is the advent of Diffusion Models and Large Language Models, which revolutionize both model formulation and the creative application of these tools.

\subsection{The Behavioral Approach: System 1 and System 2}

The list above can also be viewed from the perspective of the behavioral approach to research, inspired by the System 1-2 methodology of \cite{kahneman_thinking_2013}. According to Kahneman, there are two systems of reasoning:
\begin{itemize}
    \item System 1 reasoning is rapid, intuitive, automatic, and unconscious; it generates solutions instantaneously based on our past training.
    \item System 2 reasoning is slower, deeper, logical, and effortful. It requires our full attention and aids us in creative problem-solving.
\end{itemize}
When developing items lower on the list of AI topics above, such as Reinforcement Learning or Generative AI, we treat them as System 2 tasks, while using items higher on the list automatically, as if employing System 1.

In the following sections, we delve deeper into the System 1-2 approaches to advancing sciences with AI.

\section{Defining "My Sciences"}\label{sec:my-sciences}

It is time to introduce what I mean by "sciences," or more accurately, "my sciences" — those in which I can claim some expertise. The sciences I discuss encompass the realms of physical, social, and engineering domains.

\subsection{Physical Sciences}
I prioritize physical sciences because my background is firmly rooted in theoretical and mathematical physics, as well as applied mathematics. Within the realm of physical sciences, my primary focus centers on statistical mechanics, fluid dynamics, and, more specifically, turbulence, often referred to as statistical hydrodynamics among theorists. We will explore this topic in greater depth in the subsequent sections of the paper. Additionally, I maintain a keen interest, albeit with fewer published papers, in various other physics disciplines spanning materials science, quantum physics, astro-physics, and geo-physics.

\subsection{Social Sciences}

In the realm of social sciences, my fascination lies in the interactions of agents, be they humans, animals, or robots. I specialize in statistical modeling, ensemble analysis, and spatio-temporal resolutions, particularly when studying collective social phenomena -- such as the recent pandemic we all experienced \cite{chertkov_graphical_2021,krechetov_prediction_2022,sikaroudi_unraveling_2024}.

\subsection{Engineering Sciences}

The mention of robots above in the context of control, highlights my interest in ``theoretical engineering", which may be viewed as an element of what we discuss in the next subsection under the rubric of ``Science of AI".

On the application side of engineering sciences, I am predominantly interested in the study of energy system networks, including power \cite{bienstock_chance-constrained_2014,turitsyn_options_2011,dorfler_synchronization_2013}, natural gas \cite{chertkov_cascading_2014,zlotnik_coordinated_2016}, and district heating-cooling systems \cite{metivier_mean-field_2020,chertkov_thermal_2019,valenzuela_statistical_2023}. While working on these topics I am focusing on understanding, proposing novel approaches, and developing algorithms, including AI-based \footnote{Here are highlights of my contributions in Power System Informed AI, thus far predominantly via Deep Neural Networks (DNNs): \cite{deka_structure_2017, lokhov_online_2018, stulov_learning_2020, pagnier_physics-informed_2021, pagnier_embedding_2021, afonin_which_2021, pagnier_toward_2022, ferrando_physics-informed_2024}. A review of these efforts is forthcoming in a separate article I intend to write.}, for constructing, reinforcing, managing, and controlling these energy systems. This involves accounting for customer and operator behavioral factors and integrating the energy system seamlessly with other critical infrastructures, including transportation.

\subsection{The Science of AI}\label{sec:science-AI}

In this subsection, we explore the understanding and development of AI through innovative mathematical and statistical approaches and algorithms. It's not an exaggeration to claim that the majority, if not all, of AI innovations have their roots in applied mathematics and related disciplines, such as statistical physics. My interests and contributions to AI encompass various aspects under the broad umbrella of applied mathematics, including:
\begin{enumerate}
    \item \textbf{Statistical Inference and Learning}: I specialize in working with graphical models, employing stochastic and variational methods to enhance statistical inference and learning. This involves expressing statistical information through the use of graphs, which typically represent underlying constraints or relationships. See some of my key papers on the topic \cite{chertkov_loop_2006,lokhov_optimal_2018,chertkov_gauges_2020} and also a living book \cite{chertkov_inferlo_2024}. 
    
    \item \textbf{Stochastic Optimal Control}: This encompasses various aspects, including its data-driven counterpart, reinforcement learning in its many forms, including these most popular today which utilize neural networks for enhanced performance. I have some work in this area, mainly focusing on applications to energy systems \cite{chertkov_ensemble_2017}, and most recently to control of fat tails originating from multiplicative uncertainty \cite{chertkov_universality_2023}.
    
    \item \textbf{Diffusion Models}: Quite a lot of my recent focus is on diffusion models -- a remarkable tool in generative AI. These models are renowned for their ability to generate synthetic images based on textual prompts or stated in a bit more streamlined form -- training set of images, considered as i.i.d. samples from a probability distribution. What makes the diffusion models even more intriguing is their foundation in the clever utilization of stochastic ordinary differential equations (ODEs) and time-inversion principles, stemming from non-equilibrium statistical mechanics and the aforementioned stochastic optimal control. We will be discussing more of this in Sections \ref{sec:time-reversal},\ref{sec:diff},\ref{sec:diff-1},\ref{sec:diff-next}.
    
    \item \textbf{Transformers}: As a standout generative AI tool, transformers introduced in \cite{vaswani_attention_2017} form the foundation of recent groundbreaking advancements, including sophisticated Large Language Models such as ChatGPT. Transformers operate incrementally, extending their output word-by-word (or token-by-token, particle-by-particle) from a given input or query. Yet, at each step, the ensemble of existing particles -- comprising both the output generated and the input provided thus far -- undergoes a dynamic evolution, effectively converging on a lower-dimensional manifold or distribution that informs the sampling of the subsequent particle. As suggested by \cite{geshkovski_emergence_2023}, the impressive efficacy of transformer models can be ascribed to their function as dynamical systems, particularly how they leverage the tendency of multi-particle systems to demonstrate clustering or exhibit chaotic dynamics. My keen interest lies in exploring whether the multi-particle (Lagrangian) interpretation of turbulent flows discussed in Section \ref{sec:multi} could enlighten the design of more refined transformers and, by extension, generative models at large.
\end{enumerate}

\section{Back to the Main Questions: How Does (and Will) AI Affect Sciences?}\label{sec:AIforSc}

To comprehend the influence of AI on scientific research, we will begin with a general discussion of physical sciences. However, and as we proceed, we will illustrate our points with specific examples, such as those drawn from the field of turbulence (statistical hydrodynamics).

\subsection{Physics-Agnostic AI discovers Physics}

In the turbulence research community, AI tools have been increasingly applied, specifically {  DNNs} within supervised learning (treating turbulence as an input-output function) or generative models for handling spatial or spatio-temporal snapshots of turbulence. These AI tools have been employed to process these snapshots, treating them as if they were images of celebrities or segments from movies. This approach can be likened to a System 1 utilization of AI in turbulence research.

Remarkably, this AI approach has yielded significant insights. For instance, it has been observed that even without directly inputting physical equations, AI models have uncovered their consequences. For example, Generative Adversarial Networks (GANs) \cite{goodfellow_generative_2014}, one of the earliest tools in the generative models family of AI, have produced energy spectra in synthetic images that closely align with actual turbulence data \cite{king_deep_2018}. Furthermore, AI has unearthed intricate correlations specific to turbulence, although not all of them. In summary, in this System 1 approach to AI application, physics was not initially integrated into the process but was applied retrospectively to evaluate whether AI could faithfully preserve the physical principles.

\subsection{Exploring Physics-Informed AI}

The natural next question arises once we understand turbulence through the aforementioned System 1 approach: What should we do when the System 1 way of using AI falls short? Naturally, the community began contemplating the integration of the physics of turbulence directly into AI, giving rise to the field of \textbf{Physics-Informed AI}. Several System 2 AI approaches that incorporate physics have been proposed, and we will discuss some of them below in Sections \ref{sec:VG},\ref{sec:Eulerian},\ref{sec:multi}. However, in this section, we would like to emphasize fundamental reasons for why \textbf{Physics-Informed AI} becomes a crucial consideration:
\begin{enumerate}
    \item \textbf{Interpretation}: When our aim is not solely to be a focused prediction machine but also to gain insight into the equations, constraints, or symmetries that underlie the data. We do not necessarily require exact knowledge of these equations, constraints, or symmetries. Instead, we can introduce parameterized families of them, incorporate them into AI, and allow AI to determine not only hidden parameters (e.g., within Neural Networks) but also physical parameters. In this scenario, \textbf{Physics-Informed AI} not only provides predictions but also specific values for physical coefficients. For example, it can characterize inter-particle interaction potentials or diffusion coefficients, enabling the interpretation of underlying phenomena.

    \item \textbf{Extrapolation}: This scenario arises when we possess a limited number of samples from a specific (often rare) regime of interest, but we have an abundance of data from other regimes. In such cases, the integration of physics into AI becomes invaluable, especially when we anticipate that both distinct regimes are governed by the same or similar equations. Effective extrapolation with AI is most promising when there is confidence that the equations governing high-data and low-data regimes align or bear similarities. Additionally, shared symmetries or constraints between these regimes further enhance the effectiveness of extrapolation.

    \item \textbf{Understanding via Reduced Modeling}: Consider the aforementioned curse of dimensionality, a concern in many physical disciplines heavily reliant on computations. Even when we have an equation believed to be universal and accurate for various phenomena, such as the Navier-Stokes equation in fluid mechanics, running it in challenging cases, like at high Reynolds numbers, becomes prohibitively expensive. This is where model reduction comes into play. We postulate reduced models with significantly fewer degrees of freedom (for example changing from {  Partial Differential Equations (PDE) to Ordinary Differential Equations (ODE)}), but this introduces a multitude of options for specifying these reduced models, laden with uncertainty. Here's where physical hypotheses {  become} handy. We discuss hypotheses that can be parameterized in terms of physically meaningful parameters (as mentioned earlier). Alternatively, we can use Neural Networks and other AI tools to express degrees of freedom about which we have limited knowledge.
\end{enumerate}
Importantly, all of the above relies on data we can absolutely trust -- the so-called Ground Truth (GT) data. 

{  The term GT in this manuscript refers to the high-fidelity data used for training and validating AI models. In the context of turbulence simulations discussed in Section \ref{sec:turb}, GT typically consists of high-fidelity Direct Numerical Simulation (DNS) data. For diffusion models, discussed in Section \ref{sec:stat-mech-to-AI}, GT refers to the ensemble of real (not synthetic) images that the model aims to replicate. It is important to note that while GT data is considered a standard reference, it may contain errors due to numerical approximations or human factors. These errors can affect the training and performance of AI models. Acknowledging and modeling these errors is crucial for improving the robustness of AI methods. Traditional approaches in AI and ML, such as Graphical Models, have been employed to account for errors and uncertainties in the data (notably to de-noise images, see e.g., \cite{koller2009probabilistic}), and similar methods can be extended to modern AI techniques.}

We conclude this subsection by offering a nuanced perspective to differentiate between equations-informed and the broader concept of physics- or quantitative science-informed approaches. The distinction, elaborated upon on example of turbulence with references and a detailed discussion in Section \ref{sec:turb}, hinges on the realization that often physics transcends (or starts beyond) equations. The physics may manifest itself through conservation laws, relational constructs, or as a reduction to low-dimensional manifold representing essential degrees of freedom. An in-depth exploration of this equations-free Physics-Informed AI is presented in Section \ref{sec:Eulerian}. Furthermore, physical equations often contain phenomenological elements marked by interpretative but indeterminate parameters or by deterministic and stochastic functions of significant uncertainty. This phenomenology-based Physics-Informed AI is discussed in Sections \ref{sec:VG} and \ref{sec:multi}, illustrating how these concepts are integrated within AI methodologies.

In fact the aforementioned Physics Informed AI which is equations-free and/or phenomenology based has an interesting pre-history. The concept of an equation-free methodology traces back to at least the work of  \cite{kevrekidis_equation-free_2002}, with even earlier antecedents cited therein. This approach entails inferring equations from data, at times incorporating physics-informed biases such as symmetries or constraints, and at times not. Various notable and highly popular developments have emerged from this idea, including the ``distillation of free-form natural laws" \cite{schmidt_distilling_2009}, ``dynamic mode decomposition" \cite{kutz_dynamic_2016}, ``operator inference" \cite{peherstorfer_data-driven_2016,kramer_learning_2024}, the ``discovery of governing equations" \cite{brunton_discovering_2016}, and the ``universal linear embedding" (or Koopman embedding) \cite{mezic_spectral_2005,williams_datadriven_2015,lusch_deep_2018}. This last concept has been further enhanced to include temporal correlations through the Mori-Zwanzig formalism of statistical physics \cite{lin_data-driven_2021}.

Furthermore, the method initially described as "Artificial neural networks for solving ordinary and partial differential equations" \cite{Lagaris_nn_ode98} has undergone a resurgence as Physics-Informed Neural Networks (PINNs) \cite{raissi_physics_2017,raissi_physics_2017-1,karniadakis_physics-informed_2021}. {  It was also termed in \cite{karniadakis_physics-informed_2021}, somehow more pretentiously, as Physics-Informed Machine Learning (PIML). In the author's opinion, and given the discussion of the pre-history above, these terms -- PINN and PIML, and especially the latter one used in the literature extensively prior to the aforementioned papers, in particular to name a series of Los Alamos NL workshops on Physics Informed Machine Learning conducted in 2016, 2018, 2020, and 2022 in Santa Fe and with the next one scheduled for October of 2024 -- are much more general and inclusive, encompassing many other approaches, such as equation-free and phenomenology-guided physics-informed methods.} PINNs operate on the premise that the underlying ODEs or PDEs are precisely known, subsequently enhancing computational techniques -- historically, the collocation-point method -- by integrating the equations into the loss function for DNN training. While PINN models often yield faster results than {  DNS}, they may do so at the cost of some accuracy.

\subsection{AI to Discover New Laws of Physics}

This subsection outlines a dream, a vision yet to be realized. Consider it a daring proposal for our future work and an invitation for others to join in this ambitious endeavor.

In this context, it's crucial to embrace the spirit of bold hypothesis formation. Gone are the days of discarding seemingly unconventional ideas. With the advent of AI techniques, particularly our newfound ability to optimize in high dimensions (thanks to AD), we can adopt a Bayesian approach and consider multiple hypotheses concurrently, possibly guided by some intuitive priors.

This shift allows us to move away from subjective hypothesis selection and rely more on data-driven approaches. AI's capacity to assess and choose more scientifically reliable hypotheses based on data offers invaluable guidance. It's important to note that this approach may not provide proof but serves as an additional tool to steer our attention towards what's worth investigating and what may be disregarded. {  We briefly explore some intriguing steps in this direction in Section \ref{sec:turb-next}.}

\section{Statistical Mechanics guides development in AI}\label{sec:stat-mech-to-AI}

We demonstrate the role of statistical mechanics in informing artificial intelligence through the example of diffusion models \cite{vincent_connection_2011,sohl-dickstein_deep_2015,ho_denoising_2020}, which have revolutionized the field of AI. Remarkably, these models were constructed upon a sequence of concepts rooted in statistical mechanics and, more broadly, applied mathematics.

Our aim is to delineate the current state of the art in the field using the intuitive language of physics, particularly that of statistical physics, rather than through formal mathematical exposition.

\subsection{Markov Chain Monte Carlo}

Consider the problem of generating independent and identically distributed (i.i.d.) samples from a given probability distribution in a high-dimensional space, $p(\bm{x} \in \mathbb{R}^d)$. A viable approach involves the construction of Markov Chain Monte Carlo (MCMC) algorithms that simulate the dynamics governed by the following stochastic differential equation:
\begin{equation}\label{eq:MCMC}
\tau d\bm{x}(t) = \bm{\psi}(\bm{x}) dt + d\bm{w}(t), \quad \bm{\psi}(\bm{x}) \doteq -\nabla_{\bm{x}} \log p(\bm{x}),
\end{equation}
where $\bm{w}(t)$ represents the Wiener process with unit covariance matrix in $d$ dimensions, and $\bm{\psi}(\bm{x}) \in \mathbb{R}^d$ denotes what we will call (following {  originally stiatistics, and then} AI jargon) the score function \footnote{  The term "score function" originates from statistics, dating back to the foundational work of Ronald Fisher; see e.g., \cite{score}. It has been widely used in computational statistics, particularly in methods such as MCMC and Hamiltonian Monte Carlo. In the context of machine learning, it has been employed in \cite{hyvarinen_estimation_2005}. More recently, the role of the score function in diffusion models was clearly described in \cite{song_generative_2020}. We thank the anonymous reviewer for highlighting these important references and providing clarity on the historical context of the term.}
. Equilibrium statistical mechanics informs us that the MCMC, as defined by Eq.~(\ref{eq:MCMC}), will ultimately converge in a statistical sense. This implies that, given a set of outputs $\bm{x}(t_1), \ldots, \bm{x}(t_S)$ at successive times $t_1, \ldots, t_S$ -- each separated by at least the so-called mixing time -- the sampled data points will be i.i.d.

\subsection{Annealing}

A significant challenge in sample generation using the approach detailed in Eq.~(\ref{eq:MCMC}) is the potentially prohibitive mixing time. This leads us to the question: How can one effectively reduce the mixing time?

The concept of ``annealing'', inspired by metallurgical processes and formally introduced in statistical and computational contexts in \cite{kirkpatrick_optimization_1983}, provides an intriguing solution. Rather than relying on a statistically stationary process, annealing suggests the use of a non-stationary or non-autonomous evolution. This is achieved by substituting the stationary distribution $p(\bm{x})$ in Eq.~(\ref{eq:MCMC}) with a time-dependent distribution $p(\bm{x}|t)$. We can then gradually alter $p(\bm{x}|t)$ from an initial, easily-sampled distribution $p_0(\cdot)$ to the target distribution $p(\bm{x})$. This process hinges on two critical considerations: firstly, the initial distribution $p_0(\cdot)$ can be conveniently chosen to facilitate easy sampling; secondly, if changes to $p(\bm{x}|t)$ are sufficiently slow -- thus, adiabatic -- we can ensure a seamless transition from the simple initial distribution $p_0(\bm{x})$ to the complex target distribution $p(\bm{x})$.

The technique of annealing has been shown to be effective, offering not just empirical success but also providing a theoretical basis for constructing proofs for sampling complex distributions, as demonstrated in \cite{jerrum_approximating_1989}. However, annealing is also known for its slow convergence; the time required to produce subsequent i.i.d. samples is not only significant but also scales unfavorably with the size of the system \footnote{The issue of the big-O complexity of the annealing scheme is resolved mathematically through the so-called Fully Polynomial Randomized Approximation Schemes (FPRAS). These schemes are developed for a very special class of relatively simple models, such as attractive (ferromagnetic) Ising models. The complexity of even the best FPRASs is polynomial. For example, for certain instances of the Ising model, the time complexity can be \(O(n^2 \log \frac{1}{\epsilon})\), where \(n\) is the number of nodes in the graph and \(\epsilon\) is the approximation parameter. See \cite{jerrum_approximating_1989} for additional details.}.

\subsection{Fluctuation Theorem}

The journey toward more efficient sampling techniques has been greatly informed by recent advancements in non-equilibrium statistical mechanics, particularly those encapsulated by the ``fluctuation theorem'' and the ``Jarzynski relation" \cite{jarzynski_equilibrium_1997,kurchan_fluctuation_1998,crooks_entropy_1999}. These concepts propose that certain observables (expectations over the probability distribution $p(\bm{x})$) can be accurately reconstructed by manipulating $p(\bm{x}|t)$ at rates surpassing adiabatic conditions. This method allows for the finite-time, precise calculation of these observables, a stark contrast to the infinite time required in the asymptotic adiabatic scenario.

\subsection{Time Reversal} \label{sec:time-reversal}

While the Fluctuation Theorem provides an elegant solution for specific observables, the broader challenge of general i.i.d. sampling from $p(\bm{x})$ remains. To address this, consider a theoretical innovation with practical implications. Recalling the stochastic dynamics generalized by Eq.~(\ref{eq:MCMC}):
\begin{equation}
\label{eq:forward}
d \bm{x}(t) = \bm{f}(t;\bm{x}) dt + d\bm{W}(t),
\end{equation}
where the ``drift'' term $\bm{f}(t;\bm{x}(t))$ depends on time, and the Wiener process $\bm{W}(t)$ has a covariance matrix $\bm{\kappa}(t;\bm{x}(t))$, we arrive at the Fokker-Planck equation for $p(\bm{x}|t)$:
\begin{equation}
    \label{eq:FP}
    \left(\partial_t +\nabla_{\bm{x}} \cdot \bm{f}(t;\bm{x})\right)p(\bm{x}|t) = \kappa_{ij}(t;\bm{x})\nabla_i \nabla_j p(\bm{x}|t),
\end{equation}
assuming Ito-regularization forward in time, with $\nabla_i$ denoting $\partial_{x_i}$, and employing the Einstein summation convention for repeated indices. Considering the forward-time evolution from $t \in [0, T]$, could we theoretically invert the arrow of time? Can a backward-running stochastic process from $t \in [T, 0]$ produce the same marginal probabilities at all intermediate times $t$ as those governed by Eq.~(\ref{eq:FP})?

Remarkably, as established four decades ago in \cite{anderson_reverse-time_1982}, the answer is yes. The reverse-time stochastic process described by
\begin{equation}
\label{eq:reverse}
d \bm{x}(t) = \left(\bm{f}(t;\bm{x}) - \bm{\kappa}(t;\bm{x})\nabla_{\bm{x}} p(\bm{x}|t)\right) dt + d \tilde{\bm{W}}(t),
\end{equation}
with $\tilde{\bm{W}}(t)$ being the time-reversed Wiener process sharing the same covariance as $\bm{W}(t)$, fulfills this role. The ``drift'' term in the reverse-time Stochastic Ordinary Differential Equation (SODE) (\ref{eq:reverse}) is thus amended by a term proportional to the gradient of the Fokker-Planck (FP) Eq.~(\ref{eq:FP}) characterizing the forward-time process (\ref{eq:forward}). Verifying that the FP equation for the reverse-time SODE (\ref{eq:reverse}) aligns with the FP Eq.~(\ref{eq:FP}) for the forward-time process is a straightforward exercise. (See e.g. \cite{behjoo_u-turn_2023} for details.)

\subsection{Generating Synthetic Samples}

Having established a theoretical foundation, we now seek to translate what may appear to be a purely formal exercise into a practical sampling algorithm. We are on the cusp of achieving this — it necessitates only a slight modification to our initial approach. Suppose we possess a set of samples from $p(\bm{x})$, but the distribution $p(\bm{x})$ itself is not explicitly known; it is characterized solely by $S$ i.i.d. samples $\bm{x}^{(1)}, \ldots, \bm{x}^{(S)}$. We initiate the direct-in-time stochastic dynamics governed by Eq.~(\ref{eq:forward}) using one of these samples, say $\bm{x}^{(s)}$, and propagate it forward to time $T$, yielding $p(\bm{x}|t; \bm{x}(0) = \bm{x}^{(s)})$ for $t \in [0, T]$. By aggregating $S$ such forward-time paths into an empirical distribution, $\frac{1}{S}\sum_{s=1}^S p(\bm{x}|t; \bm{x}(0) = \bm{x}^{(s)})$, we can approximate $p(\bm{x}|t)$ for use in Eq.~(\ref{eq:reverse}). This reconstructed distribution allows us to then simulate the reverse-time process and generate novel synthetic samples from $p(\bm{x})$, thereby fulfilling our objective.

\begin{figure}[h!]
    \centering
    \includegraphics[width=\textwidth]{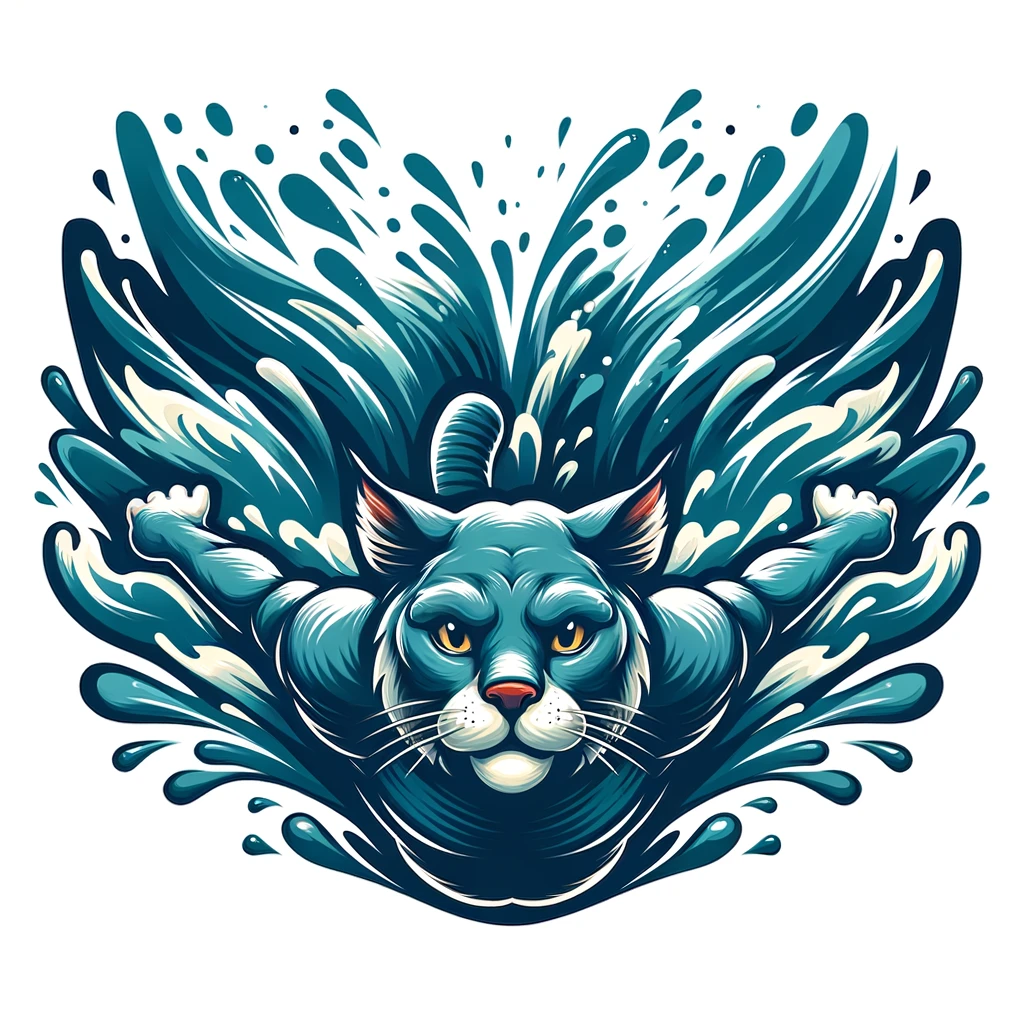}
    \caption{GPT-4 prompt: Illustration of the University of Arizona's bobcat mascot swimming butterfly style in a turbulent river. {  (The entire prompt was used to generate the image without providing additional information such as color, symmetry, or viewing direction.)}}
    \label{fig:swimming}
\end{figure}
The methodology outlined here mirrors the diffusion algorithms integral to systems like ChatGPT, which are capable of generating imagery from textual descriptions, such as the depiction shown in Figure~\ref{fig:swimming}.

How do the diffusion models and algorithms we have outlined contribute to the algorithms employed by systems like ChatGPT \cite{OpenAI_ChatGPT} to synthesize new images from textual prompts? Diffusion models operate by incrementally introducing noise into the data, denoted by $\bm{x}(0) = \bm{x}^{(s)}$, where in the context of images, $\bm{x}(0)$ represents the vector form of an image, with each component corresponding to a pixel and its RGB color values. Hence, the dimensionality of a color image, segmented into $128 \times 128$ pixels, is $128 \times 128 \times 3$. This noise addition follows a discretized interpretation of the forward-time stochastic ODE (\ref{eq:forward}), leading to a state at time $T$ where the image's original data $\bm{x}(T)$ is completely obfuscated. The algorithm is then trained to invert this noising process, commencing from the noise-induced state $\bm{x}(T)$ and progressively denoising it, utilizing a time-discretized counterpart of the reverse-time stochastic ODE (\ref{eq:reverse}) to regenerate fresh data samples at $\bm{x}(0)$. By conditioning the reverse process on particular inputs -- like textual prompts in image generation \cite{ramesh_hierarchical_2022} -- the algorithm produces outputs that are congruent with the given prompts.

\subsection{Diffusion Algorithm}\label{sec:diff}

Let's consider the practical implementation of the fundamental diffusion algorithm, which aims to generate new unbiased i.i.d. samples without any prompting or bias. A key advantage of this algorithm is its remarkable universality; it performs well even with significantly simplified assumptions for drift and diffusion. A common practical approach involves approximating Eq.~(\ref{eq:forward}) with the Ornstein–Uhlenbeck (OU) process:
\begin{equation}
    \label{eq:OU}
    d{\bm x}(t)=-\frac{1}{2}\beta(t){\bm x}(t)dt+\sqrt{\beta(t)}d{\bm w}(t),\quad {\bm x}(0)\sim \left\{{\bm x}^{(s)} \mid s=1,\ldots,S\right\},
\end{equation}
which yields the marginal probability as a Gaussian/normal mixture over the GT samples:
\begin{equation}
    \label{eq:GM}
    p({\bm x}|t)=\frac{1}{S}\sum_{s=1}^S\mathcal{N}\left({\bm x} \Bigg| e^{-\frac{1}{2}\int_0^t \beta(t')dt'} {\bm x}^{(s)}, \left(1 - e^{-\int_0^t \beta(t')dt'}\right){\bm I}\right).
\end{equation}
This expression enables the replacement of stochastic path simulations with direct computations, using the GT samples to calculate the respective means and covariances.

The subsequent crucial step involves "learning" the drift for the reverse of Eq.~(\ref{eq:OU}):
\begin{equation}
    \label{eq:OU-reverse}
    d{\bm x}(t) = -\beta(t)\left(\frac{{\bm x}(t)}{2} + {\bm \psi}(t;{\bm x}(t))\right)dt + \sqrt{\beta(t)}d\tilde{\bm w}(t),\quad {\bm x}(T)\sim \mathcal{N}(0,{\bm I}),
\end{equation}
with the score function ${\bm \psi}(t;{\bm x}(t))$ defined as $-\nabla_{\bm x}\log p({\bm x}|t)$. Notably, the reverse process (\ref{eq:OU-reverse}) is inherently nonlinear, necessitating actual simulations.

Herein lies the transformative power of AI: we approximate the score function using a {  DNN} parameterized by ${\bm \theta}$. The training process involves optimizing the discrepancy between the true score function and its NN approximation, evaluated over a uniform temporal distribution within $[0,T]$ and spatial distribution according to Eq.~(\ref{eq:GM}):
\begin{equation}
    \label{eq:training}
    \min_{\bm \theta} \mathbb{E}_{\substack{t\sim \text{U}(0,T) \\ {\bm x}(0)\sim\{{\bm x}^{(s)}\} \\ {\bm x}(t)\sim p({\bm x}|t)}}\left[\| {\bm \psi}(t;{\bm x}(t)) - \text{NN}_{\bm\theta}(t;{\bm x}(t))\|^2\right].
\end{equation}

With this training completed, we can now generate new samples from the distribution represented by its GT samples. We simply initialize the reverse process (\ref{eq:OU-reverse}) with a draw from the univariate normal distribution ${\bm x}(T)\sim\mathcal{N}(0,{\bm I})$ and consider ${\bm x}(0)$ as the new synthetic sample. Remarkably, although these new samples are distinct, they bear a "similarity" to the GT samples -- a testament to the {  DNN}'s capacity to effectively smooth the distribution around $t=0$ (where the empirical distribution turns simply into a sum of the $\delta$-functions).

\subsection{The Dawn of AI Innovation: Explorations and Enhancements}\label{sec:diff-1}

As we navigate through the nascent stages of the AI revolution, it's evident that the integration of applied mathematics and statistical mechanics will play a pivotal role in propelling the field forward. Although we are just at the beginning, there are notable advancements within the outlined general framework that merit attention, even if explored in a preliminary manner. These developments signify the untapped potential waiting to be harnessed, illustrating the fertile ground for innovation that lies at the intersection of AI, mathematics, and physics.

\subsubsection{The U-Turn Technique}

\begin{comment}
\begin{figure}[h!]
\centerline{\includegraphics[scale=0.5]
	{figs/image_loop_cosine.jpg}} 
	\caption{
 Synthetic images generated from the model presented in \cite{behjoo_u-turn_2023}, starting at $t=0$, and executing U-turns at various times within the $[0, 1000]$ interval. The model was trained on a set of 1000 ground truth (GT) butterfly images sourced from the Hugging Face Hub (\url{https://huggingface.co/datasets/huggan/smithsonian_butterflies_subset}). Early U-turns, close to $t=0$, yield images resembling the original GT images. Conversely, U-turns performed later, specifically at $t \gtrsim 600$, result in the creation of novel synthetic images.
 }
	\label{fig:loopy_dynamics} 
\end{figure}

The U-turn approach, as introduced in \cite{behjoo_u-turn_2023}, explores the pivotal insight that the core measure of the diffusion models' efficiency lies in their ability to rapidly attenuate correlations during the reverse or de-noising phase. Contrary to extending the forward diffusion to its full extent, the U-turn technique advocates for an earlier reversal - illustrated in Fig.~(\ref{fig:loopy_dynamics}). Specifically, it employs a shortened forward diffusion phase, immediately followed by the customary reverse dynamics, which commence from the terminal state of the forward phase. To calibrate the optimal timing for the U-turn, the methodology employs a suite of analytical tools, including auto-correlation analysis, the weighted norm of the score function, and the Kolmogorov-Smirnov test for Gaussianity. 
\end{comment}

{ 

To illustrate the forward and reverse processes, and their use in training and generating new samples, we provide an example introduced in \cite{behjoo_u-turn_2023} and called the U-turn diffusion model. This model explores diffusion processes consisting of a forward noise-injecting stage and a reverse de-noising stage to encode information about the GT samples.

The U-turn diffusion model modifies the traditional approach by shortening both the forward and reverse processes, starting the reverse dynamics from the final configuration of the forward process. This modification aims to optimize the generation of synthetic samples that are independent and identically distributed (i.i.d.) according to the probability distribution represented by the GT samples.

In the experiments of \cite{behjoo_u-turn_2023} with the ImageNet-64 dataset \cite{imagenet}, the U-turn diffusion model achieved state-of-the-art Fréchet Inception Distance  scores with fewer Neural Function Evaluations. Notably, this approach provided a 1.35-fold speed-up in inference without requiring retraining.

Figure~\ref{fig:uturn_example} illustrates the basic Score Based Diffusion process and the U-turn concept. The analysis reported in \cite{behjoo_u-turn_2023}, based on new tests such as the U-Turn Auto-Correlation function test, revealed that making the U-turn earlier is beneficial but should not be done too early to avoid memorization of the original image. These tests guide the optimal timing for the U-turn, ensuring the generation of high-quality synthetic images.

\begin{figure}[h]
    \centering
    \includegraphics[width=\textwidth]{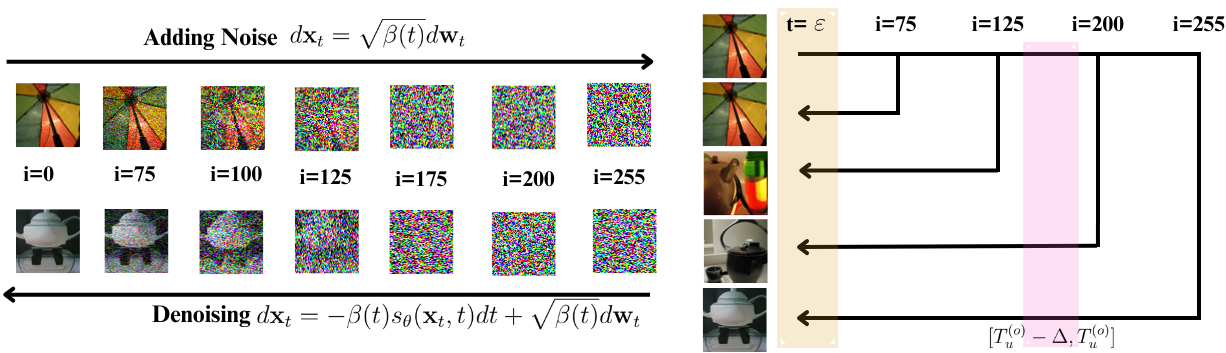}
    \caption{Illustration, from \cite{behjoo_u-turn_2023}, of the basic Score Based Diffusion (left) and U-turn (right) concepts. The U-turn involves an earlier transition from the forward to the reverse process, optimizing sample generation.}
    \label{fig:uturn_example}
\end{figure}

}

\subsubsection{Bridge-Diffusion Techniques}

The bridge-diffusion methodology offers a compelling variation to the conventional time reversal mechanisms in diffusion models. Unlike traditional approaches that aim to align the marginal probabilities across auxiliary times with those of the original process through time reversal, the bridge-diffusion strategy modifies the foundational process by anchoring it to a predetermined outcome at a specified final moment, notably at $T=1$. This technique, drawing from Doob's h-transform known in stochastic process theory, facilitates a transition from an easily sampleable distribution to one modeled on real data samples, as elucidated in \cite{peluchetti_non-denoising_2021}. A critical advantage of the bridge-diffusion model over the standard denoising diffusion paradigm lies in its elimination of uncertainties regarding the final time $T$. While traditional models necessitate extending $T$ towards infinity to ensure theoretical perfection, the bridge-diffusion model achieves this precision by designating $T$ to a finite, predetermined value, such as $T=1$.

\subsubsection{Space-Time Diffusion}

\begin{figure}[h!]
\centerline{\includegraphics[scale=0.7]
	{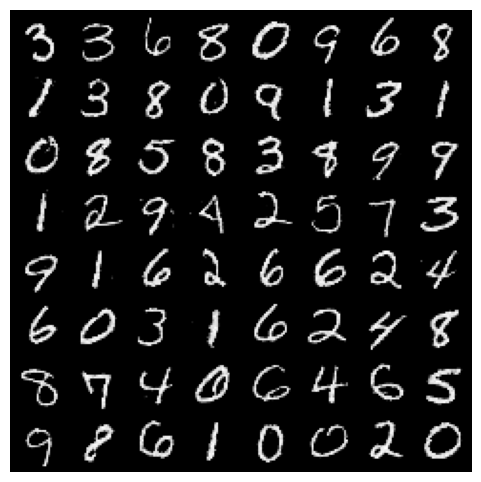}} 
	\caption{This visualization showcases generative capability of the space-time bridge-diffusion model from \cite{behjoo_space-time_2024} to synthesize high-fidelity images. The diffusion model was trained on the MNIST dataset, an established benchmark in the machine learning domain for handwritten digit recognition, which consists of $70,000$ images with dimensions of $28\times 28$ pixels.}
    \label{fig:bridge-mnist}
\end{figure}

The foundational stochastic processes underpinning the diffusion algorithms previously outlined exhibit limited mixing among the $d$ components of a sample. To address this shortcoming, the incorporation of space-time mixing strategies into the foundational stochastic process has been proposed, aiming to enhance diffusion across both temporal and spatial dimensions. This approach, delineated in \cite{behjoo_space-time_2024}, builds on three synergistic stochastic processes engineered to facilitate efficient transport from a simple initial probability distribution to the complex target distribution reflected in the ground truth (GT) samples. These processes include: (a) linear processes integrating space-time mixing to produce Gaussian conditional probability densities, (b) bridge-diffusion counterparts conditioned on specific initial and final states, and (c) nonlinear stochastic processes optimized via score-matching techniques. The model training phase involves calibration/optimization of both the nonlinear and, if necessary, linear models to ensure a high degree of fidelity to the GT data. 
Refer to Fig.~\ref{fig:bridge-mnist} for a visual illustration.

\subsubsection{Phase Transitions in Optimally Trained Diffusion Models}

In the recent work \cite{biroli_dynamical_2024} a detailed analysis inspired by the concept of phase transitions in physics is applied to optimally trained diffusion models in the limits of infinite dimensions and sample sizes $d, S \to \infty$. This study commences with an analytical exploration based on large-deviation (saddle-point) analysis, focusing on sampling from a synthetic distribution characterized by a sum of two Gaussian distributions. Subsequent experimental validations on a pre-trained diffusion model for realistic high-dimensional data set of images corroborated the theoretical insights, unveiling three distinct dynamical regimes that emerge along the reverse generative path of the diffusion process.

These regimes include a 'speciation' phase, where the intrinsic structure of the dataset begins to manifest, followed by a 'collapse' phase, during which the model's trajectory converges towards previously learned data points, showcasing attraction towards memorized information. This nuanced understanding of diffusion model dynamics, particularly in relation to the dynamic phase transitions observed in statistical physics, offers invaluable insights for operational optimization.

In light of the discussion above on strategic reversals (U-turns \cite{behjoo_u-turn_2023}) in the training of diffusion models, the insights derived from the phase-transition analysis are particularly enlightening. It is suggested that implementing a U-turn beyond the 'collapse' phase could optimize the model's performance by leveraging the identified dynamic regimes. Further development of this statistical physics prospective may allow for a more informed decision-making process regarding the timing of U-turns, potentially enhancing the model's ability to generate novel and diverse outputs while retaining high fidelity to the training data.

\subsection{Applied Mathematics (and Theoretical/Statistical Physics) for AI: What is Next?}\label{sec:diff-next}

Exciting and untapped opportunities abound in the development of applied mathematics for AI. I would like to highlight areas of particular interest (for the author, and hopefully readers):
\begin{itemize}
\item {\bf Convergence of AI Methodologies: Unifying Mathematics across Reinforcement Learning, Diffusion, and Transformers}: Reinforcement learning, diffusion, and transformers -- while traditionally seen as distinct pillars within the field of AI -- actually converge on a shared mathematical foundation. Intriguingly, these methodologies are all bound by differential equations that are not only first-order in computational time but can extend to higher orders within the multidimensional expanses of their respective state-spaces. This common ground extends to control theory formulations that guide the analysis and/or training of these diverse approaches \cite{tzen_theoretical_2019,chen_stochastic_2021,grohs_dynamical_2022}. From a statistical physics viewpoint, these AI models are subject to dynamical phase transitions \cite{geshkovski_mathematical_2024, biroli_dynamical_2024}, in the asymptotic limits of large spatial dimensions or expansive sample sizes. Such transitions can manifest significant changes in model performance, hinting at underlying commonalities in their operational mechanics.

Despite these shared mathematical principles, diffusion models, transformers, and reinforcement learning algorithms also showcase distinct attributes that set them apart, especially in their applications. Transformers and reinforcement learning, for instance, are predominantly applied to tasks characterized by an ordered state space, a concept central to many natural language processing challenges. This order is mirrored in the sequential processing inherent to transformers and the iterative state-action-reward loops in reinforcement learning. In contrast, diffusion models have demonstrated their prowess in handling the stochastic generation of two-dimensional -- and potentially even higher-dimensional -- image spaces.

The insights gained from understanding these similarities and differences are invaluable. They pave the way for the development of new, standalone AI techniques that are specifically tailored to tackle challenging tasks that remain elusive to current methodologies. Moreover, by dissecting and synthesizing the strengths of each approach, we can refine and enhance existing techniques. Perhaps most exciting is the potential for creating hybrid models -- innovative fusions that draw from each method's unique advantages to transcend the capabilities of any single approach.

\item {\bf Empirical Success and the Complexity Paradox of AI: Navigating through Typical Cases}: The empirical success of AI, often lacking full theoretical understanding, necessitates extensive fine-tuning of loss functions and hyperparameters to make optimization tractable. Remarkably, this fine-tuning leads to scaling that is linear or quasi-linear with problem size, despite complexity theory suggesting that generic problems are expected to be intractable -- have exponential complexity. This apparent contradiction is likely resolved by the fact that AI optimization typically evades worst-case scenarios posited by complexity theory, focusing instead on more typical cases. This conclusion is reminiscent of one reached in statistical mechanics, where phase transition analysis and avoidance of algorithmically challenging "glassy" phases are based on typical rather than worst-case scenarios \cite{mezard_analytic_2002}. Empirically, successful AI models seem to sidestep such glassy phases, indicating a need to adapt these concepts of typical complexity and phase transitions from statistical mechanics to the analysis and design of AI algorithms -- a formidable yet promising endeavor.
  
\item  {\bf Data Scarcity, Instantons and Importance Sampling for AI}: The abundance of data stands as a foundational pillar in the realm of Artificial Intelligence, propelling its advancement and shaping its evolution. Yet, it's crucial to acknowledge the inherent limitations of AI systems, particularly in scenarios marked by a scarcity of data points or the complete absence of Ground Truth (GT) benchmarks. Such constraints become pronounced barriers in the quest for reliable extrapolation and generalization within data-sparse regimes. While AI encounters obstacles in these contexts, the rich toolkit of applied mathematics and theoretical physics -- boasting methods like sensitivity analysis and uncertainty quantification \cite{smith_uncertainty_2013}, instantons \cite{chertkov_instanton_1997,e_minimum_2004,rotskoff_active_2021}, and importance sampling \cite{owen_importance_2019,paananen_implicitly_2021} -- offers a beacon of hope. The strategic adaptation and refinement of these more traditional statistical mechanics techniques to the nuanced demands of AI represent a research trajectory teeming with potential. This fusion of classic analytical strategies with modern computational paradigms could unlock unprecedented capabilities in AI, equipping it to navigate and thrive in the vast unknowns of data-deficient landscapes. (For an insightful exploration of how reinforcement learning (RL) and energy-based AI models are integrated into the sampling and estimation of low-probability events, see \cite{rose_reinforcement_2021} and \cite{friedli_energy-based_2023}.)

\item {\bf State-Space Structures in Diffusion Models: Harnessing PDEs and Graph Theory for Advanced Capabilities}: The diffusion models of AI, often employed in the rendering of two-dimensional imagery (and potentially high-dimensional snapshots/images), have largely not exploited the intricate structure of state-space. Such structures, implicit in the juxtaposition of image pixels or the manifold of higher-dimensional data, beckon for a more sophisticated incorporation into the AI models. It is reasonable to prognosticate that the field would undergo a significant transformation with the systematic integration of these state-space structures, particularly through the lens of PDEs that embrace both spatial and temporal dimensions (d+1), as well as through the application of graph theory and graphical models \cite{wainwright_graphical_2007,chertkov_inferlo_2024}, expressing relations (deterministic and stochastic) of the underlying applications. These mathematical structures, when finely tuned and interwoven with diffusion processes, hold the potential to significantly expand the capabilities of diffusion models, advancing them into previously uncharted and complex application realms.
\end{itemize}
We foresee that these advances will not only broaden our comprehension and enhance our proficiency in artificial intelligence but also propel substantial breakthroughs across various scientific domains where AI is pertinent -- thereby seamlessly connecting to the topic of the next section.

\section{AI guides development in Stochastic Hydrodynamics}
\label{sec:turb}

Science evolves in bursts. The author recalls one such burst in the development of statistical hydrodynamics that occurred in the late '90s. A key achievement of this period was the establishment of the origin of anomalous scaling in the so-called passive scalar turbulence, which describes the dispersal of a passive substance (like ink or a pollutant) in space and time when injected at a large scale and stirred by a multi-scale stochastic velocity field \cite{shraiman_lagrangian_1994,gawedzki_anomalous_1995,chertkov_normal_1995}. This leap forward was primarily facilitated by the linearity of the PDE that governs the dynamics of the passive scalar then  permitting the correlation functions of the scalar of an $n$-th order to be expressed in terms of the stochastic dynamics of $n$ Lagrangian particles. At that time, 25 years ago, there was also an attempt to leverage the success of this "linear" theory for its famous "nonlinear" counterpart -- turbulence in the Navier-Stokes equation \cite{chertkov_lagrangian_1999}. The proposed tetrad-model phenomenology conjectured that correlations of the velocity field could be related through a Lagrangian closure to the dynamics of four particles -- a tetrad -- providing a minimal statistical Lagrangian representation of a fluid element. The narrative of the Lagrangian closure has evolved since then, primarily along the path of validating the phenomenological details at the viscous scale, where the tetrad geometry simplifies to that of a single particle.

In what follows, we will discuss how this Lagrangian phenomenology has recently been augmented with AI \cite{tian_physics-informed_2021,woodward_physics-informed_2023,tian_lagrangian_2023}-- thus providing a prime example of how AI is propelling the advancement of the science phenomenologies from the pre-AI era.

\subsection{Scalar Turbulence: From Fields (Eulerian) to Particles (Lagrangian)}

\begin{figure}[H]
    \centering
    \includegraphics[width=0.7\textwidth]{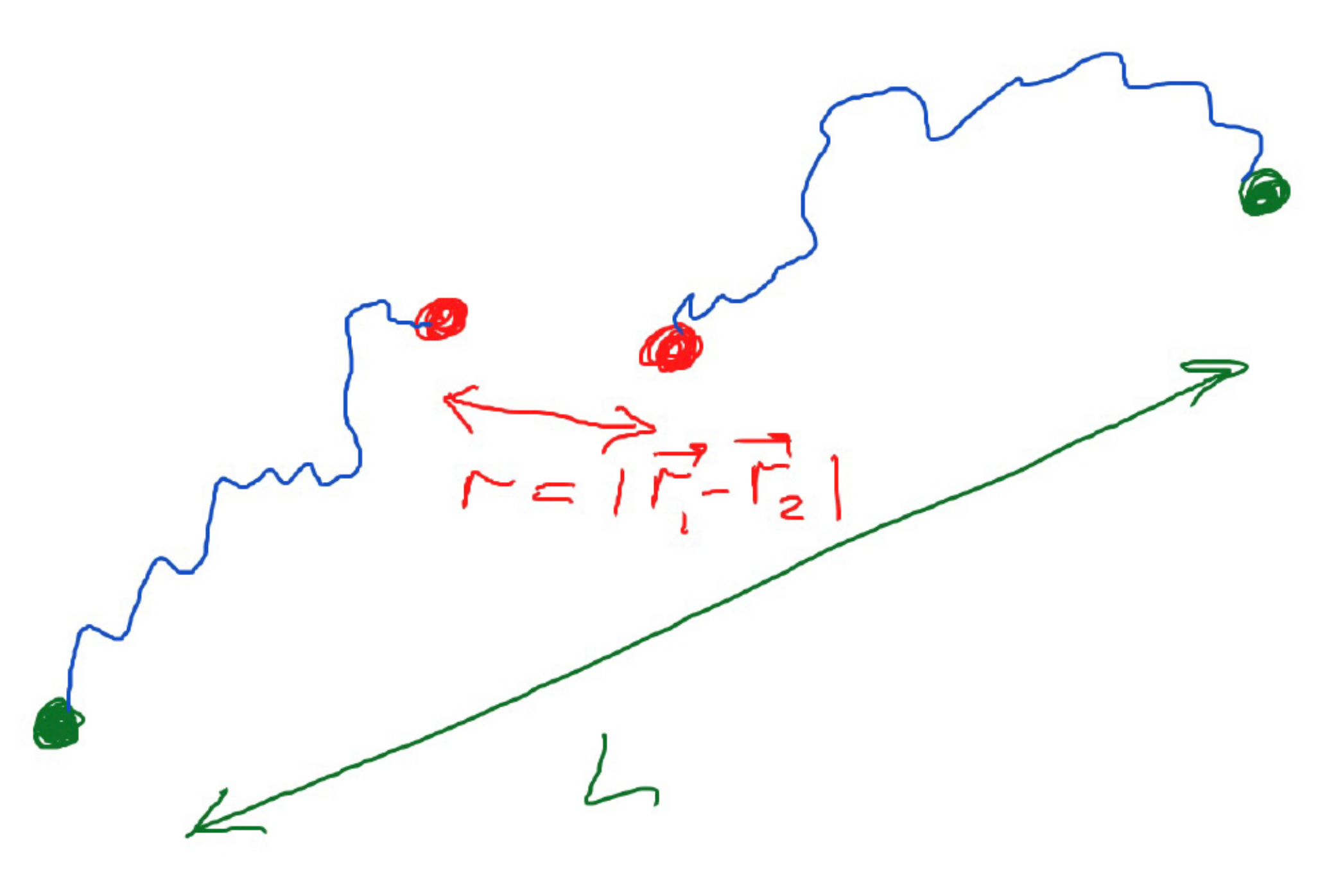}
    \caption{Schematic illustration of Lagrangian trajectories of a pair of particles relevant for computations of passive scalar correlations in Eq.~(\ref{eq:theta-2}).}
    \label{fig:scalar}
\end{figure}

The advection of the passive scalar field, $\theta(t;{\bm r})$, which evolves in space ${\bm r}\in \mathbb{R}^d$ and time $t$, is governed by the following PDE \cite{kraichnan_small-scale_1968,batchelor_small-scale_1959}
\begin{equation}\label{eq:theta-eq}
\partial_t \theta+({\bm v}(t;{\bm r})\cdot \nabla_{\bm r})\theta=\kappa\nabla_{\bm r}^2\theta+\phi(t;{\bm r}),
\end{equation}
where the velocity field, ${\bm v}(t;{\bm r})$, is an exogenously prescribed, stochastic, stationary, and spatio-temporally multi-scale field; $\phi(t;{\bm r})$ is another exogenously prescribed zero-mean stochastic field, which is independent of the velocity field, representing the injection of the scalar.

Let us demonstrate the transformation from the Eulerian (field) representation to the Lagrangian (particle) representation for an $n$-th order correlation function of the scalar by examining the example of a simultaneous correlation function of the scalar at two distinct locations, ${\bm r}_{1,2}\in \mathbb{R}^d$:
\begin{equation}\label{eq:theta-2}
    \mathbb{E}_{{\bm v};{\bm \phi}}\left[\theta(t;{\bm r}_1)\theta(t;{\bm r}_2)\right]=\mathbb{E}_{{\bm v};{\bm W}}\left[\int\limits_{-\infty}^t dt' \chi\left({\bm \rho}(t';{\bm r}_1)-{\bm \rho}(t';{\bm r}_2)\right)\right],
\end{equation}
assuming, without loss of generality, that the pair correlation function of the injection field is $\delta$-correlated in time and spatially homogeneous. This implies $\mathbb{E}_{\bm \phi}\left[\phi(t_1;{\bm r}_1)\phi(t_2;{\bm r}_2)\right]=\delta(t_1-t_2)\chi(|{\bm r}_1-{\bm r}_2|)$, where the spatial correlations encoded in $\chi(r)$ decay to zero sufficiently fast beyond the correlation scale, $L$, i.e., for $r > L$. In Eq.~(\ref{eq:theta-2}), ${\bm \rho}(t';{\bm r})$ delineates the stochastic dynamics of a Lagrangian particle, subjected to advection/drift by the velocity field and to diffusion by the Wiener process of strength $\kappa$, conditioned on arriving at time $t$ at ${\bm \rho}(t;{\bm r})={\bm r}$:
\begin{equation}\label{eq:rho-eq}
\forall i,j=1,\cdots,d,\ \forall t'\in ]-\infty;t]:\ d\rho_i(t';{\bm r})=v_i(t';{\bm \rho}(t';{\bm r}))dt'+\sqrt{2\kappa} d W_i(t'),\ \mathbb{E}\left[d W_i(t') d W_j(t')\right]=\delta_{ij} dt'.\  
\end{equation}
These formulas, as depicted in Fig.~(\ref{fig:scalar}), imply that the pair correlation function of the scalar, Eq.~(\ref{eq:theta-2}), can be estimated as $\chi(0)\mathbb{E}_{{\bm v},{\bm W}}[T(L\to |{\bm r}_1-{\bm r}_2|)]$, where $T(L\to |{\bm r}_1-{\bm r}_2|)$ denotes the time it takes
or two Lagrangian particles, initially separated at time $t-T(L\to |{\bm r}_1-{\bm r}_2|)$ by scale $L$ or less, to reach a separation of $|{\bm r}_1-{\bm r}_2|$ (assumed much smaller than $L$) at time $t$. It is noteworthy that the Lagrangian trajectories contributing predominantly to the pair correlation function of the scalar are atypical -- in a turbulent flow, the distance between a pair of particles would usually increase over time, not decrease.

This shift from the Eulerian perspective, which focuses on the passive scalar field, to the Lagrangian framework, which considers the motion of particles within the flow, has been pivotal in understanding the anomalous scaling and intermittency of scalar turbulence. For further reading, we direct the interested reader to seminal works \cite{shraiman_lagrangian_1994,gawedzki_anomalous_1995,chertkov_normal_1995} and to the review \cite{falkovich_particles_2001}.

The insights gained have also spurred efforts to adapt the Eulerian-to-Lagrangian mapping to the significantly more complex context of Navier-Stokes turbulence, which we discuss next.

\subsection{Tetrad Model: Lagrangian Phenomenology}

Consider homogeneous, isotropic turbulence in three dimensions, $d=3$, governed by the Navier Stocks equations for the velocity field, ${\bm v}(t;{\bm r})$:
\begin{equation}\label{eq:NS}
\partial_t {\bm v}+({\bm v}\cdot \nabla_{\bm r}){\bm v}=-\nabla_{\bm r} p+\nu\nabla_{\bm r}^2{\bm v}+{\bm f}(t;{\bm r}),    
\end{equation}
where the pressure field is reconstructed from the incompressibility condition on the velocity, $\nabla_{\bm r} {\bm v}=0$. The turbulence is driven the exogenous source term injecting energy at the sufficiently large scale, $L_v$, and it cascades down to the much smaller viscous (also called Kolmogorov) scale, $\nu$, where the energy is dissipated by the viscosity ($\nu$-dependent term in Eq.~(\ref{eq:NS})). Large value of $L/\nu$ indicates that the turbulence is developed (respective dimensionless quantity, called Reynolds number, is large).

It is instructive to apply the gradient operator to the NS Eq.~(\ref{eq:NS}), thus arriving at the equation for the Velocity Gradient Tensor/matrix (VGT), with the components $m_{ij}=\nabla_j v_i$
\begin{equation}\label{eq:VG}
\frac{d}{dt} {\bm m}+{\bm m}^2-{\bm I}\frac{\text{tr}[{\bm m}^2]}{3}=-{\bm h}+ \text{injection} +\text{dissipation},    
\end{equation}
where ${\bm I}$ is the $(3\times 3)$ unit matrix and ${\bm h}$ is the non-local part of the Pressure Hessian (PH) tensor/matrix with the components, $h_{ij}=\nabla_i\nabla_j p$. Notice that in transitioning from the NS Eq.~(\ref{eq:NS}) to the VGT Eq.~(\ref{eq:VG}) we have utilized the Lagrangian framework, thus replacing $\partial_t+({\bm v}\cdot \nabla_{\bm r})$ by $d/dt$, according to Eq.~(\ref{eq:rho-eq}). 

\begin{figure}[H]
    \centering
    \includegraphics[width=0.7\textwidth]{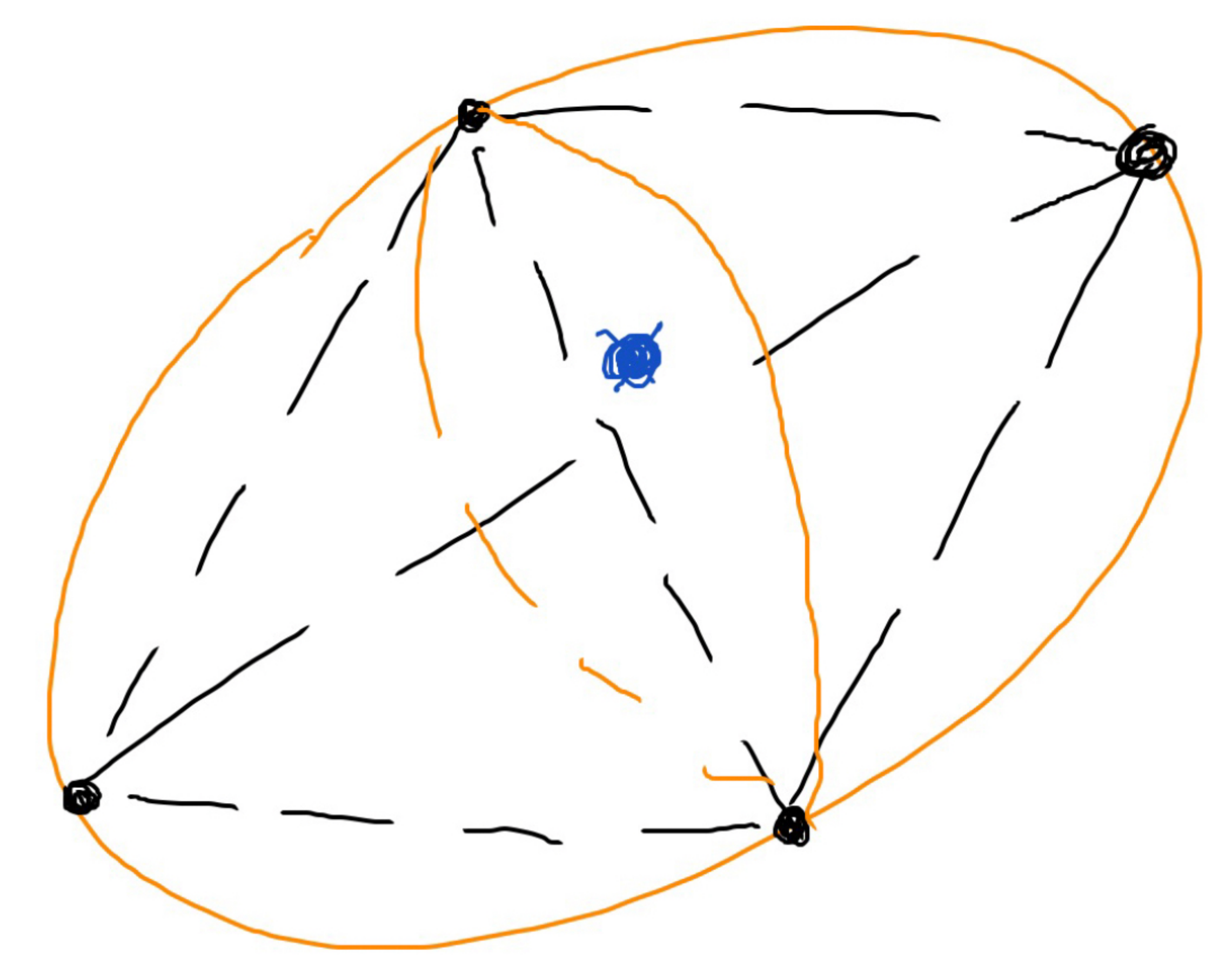}
    \caption{Schematic illustration of the minimal representation of a blob via a tetrad (four black dots/particles) or equivalently the (orange) ellipsoid. Geometry of the tetrad/ellipsoid, described in terms of the ellipsoid's tensor of inertia, ${\bm g}$, is evolving in time (not shown). (The blue dot marks the tetrad center of inertia.) }
    \label{fig:tetrad}
\end{figure}

Note that the Pressure Hessian (PH) depends on the dynamics at large scales, thus leaving Eq.~(\ref{eq:VG}) unclosed. To address this problem the approach taken in \cite{chertkov_lagrangian_1999} consisted in closing it by coarse-graining. Specifically, we introduce $\bm{M} = \int_{\text{blob}}  \bm{m}/\text{volume of the blob}$, which represents the Velocity Gradient Tensor (VGT) coarse-grained over a fluid blob with linear dimensions within the inertial range. This blob is minimally represented by a tetrad, or equivalently, an ellipsoid as depicted in Fig.~(\ref{fig:tetrad}). The result is a phenomenological framework -- that is one not derived but postulated -- suggesting the following system equations for co-dependent ellipsoid's tensor of inertia $\bm{g}$ and the velocity gradient coarse-grained (averaged) over the ellipsoid:
\begin{align}\label{eq:M}
    &d\bm{M} + (1 - \alpha)\left(\bm{M}^2 - \bm{g}^{-1}\frac{\text{tr}[\bm{m}^2]}{\text{tr}[\bm{g}^{-1}]}\right) dt = d\bm{W}_M(t), \\ \label{eq:g}
    &d\bm{g} - \left(\bm{M}^\top\bm{g} + \bm{g}\bm{M}\right)dt = d\bm{W}_g(t),
\end{align}
where $\bm{W}_M(t)$ and $\bm{W}_g(t)$ are zero-mean stochastic terms modeled as Wiener processes, with their covariances dependent on $\bm{M}$ and $\bm{g}$. The covariance matrices and the tuning parameter $\alpha$ in Eqs.~(\ref{eq:M},\ref{eq:g}) were empirically calibrated using post-factum Eulerian {  DNS}, as detailed in \cite{chertkov_lagrangian_1999}. Further calibration of the tetrad phenomenology against Lagrangian data was performed in \cite{pumir_geometry_2000}.

At this stage in the story it seems appropriate to make a step back and describe the rich pre-history which has led to development of the tetrad model~(\ref{eq:M},\ref{eq:g}). Indeed, exploration of the Velocity Gradient Tensor (VGT) within the realm of developed incompressible turbulence has been a subject of enduring interest and investigation since the early 1980s. The early closure attempts of the NS Eq.~(\ref{eq:NS}), first by \cite{vieillefosse_local_1982} and then by \cite{cantwell_exact_1992} led to the Restricted Euler (RE) equation -- which is just Eq.~(\ref{eq:VG}) with zero right hand side. However, this, so-called {  Vieillefosse} equation was clearly problematic due to its nonphysical finite time singularity. Concurrently, and leading up to \cite{chertkov_lagrangian_1999}, \cite{girimaji_diffusion_1990} embarked on exploring related stochastic models and phenomenology for the VGT. These efforts, focused on the local dynamics of the VGT (rather than its coarse-grained behavior which came later in \cite{chertkov_lagrangian_1999}) introduced stochastic modifications to the right-hand side of Eq.~(\ref{eq:VG}). The stochastic terms were interpreted as immitating effect of larger scales on smaller scales. 

The VGT methodology continued to involve since early 2000 -- \cite{pumir_geometry_2000,jeong_velocity-gradient_2003,chevillard_lagrangian_2006,chevillard_modeling_2008,wilczek_pressure_2014,johnson_closure_2016,pereira_dissipative_2016,leppin_capturing_2020,yang_dynamics_2023,tian_lagrangian_2023,das_data-driven_2023} -- with some papers in the list focusing on the statistics of the VGT itself (local VGT) while others extending it to the coarse-grained VGT. For example, and closing the discussion of the coarse-grained modeling, \cite{yang_dynamics_2023} suggested a data-driven model for the "perceived" VGT using a tetrad of four particles. One takeaway from this study was that while the tetrad model, as initially proposed in \cite{chertkov_lagrangian_1999}, aptly describes the forward evolution of the perceived VG tensor at the shortest timescales (below the Kolmogorov time scale), it struggles with longer timescales.

On the other hand, it also became clear from the studies that local closures -- that are these confined primarily to the viscous or even smaller scales -- which we continue to discuss in the next subsection -- have a merit on their on, independently on the  closures  at the larger scale where we should necessarily need to co-evolve the coarse-grained velocity gradient with the tensor of inertia. 

\subsection{Pressure Hessian as a Function of Velocity Gradient: Lagrangian Neural Closure}\label{sec:VG}

Pressure Hessian contributing the right hand side of Eq.~(\ref{eq:VG}) is a convolution of the VGT with the inverse of the Laplacian operator over the entire spatial domain
\begin{equation}\label{eq:h-m}
    h_{ij}(t;{\bm r})=-\int d {\bm r}'\frac{r'^2\delta_{ij}-r'_i r'_j}{4\pi r^5} \text{tr}\left[\left({\bm m}({\bm r}+{\bm r}')^2\right)\right].
\end{equation}
The spatially nonlocal nature of the relation between ${\bm H}$ and ${\bm m}$ is obvious here. However, assuming that contributions to the integral on the right hand side of Eq.~(\ref{eq:h-m}) from the parts of the domain which are sufficiently far from the point of observation, ${\bm r}$, average out, it seems reasonable to build a {\bf closure model for the pressure Hessian} -- to approximate ${\bm H}(t;{\bm r})$ by a function of ${\bm m}(t;r)$ evaluated at the same spatio-temporal location. However, a question arises: given the local approximation hypothesis, how can we approximate the function?

\begin{figure}[h!]
	\centerline{\includegraphics[width=3in]{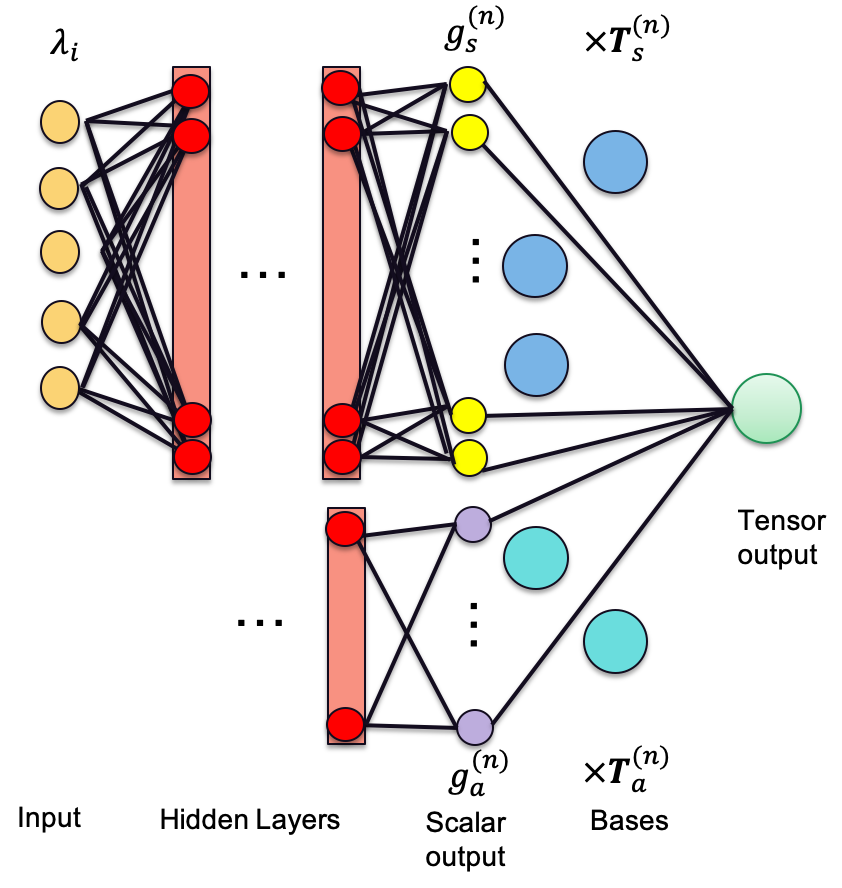}}
	\caption{The architecture of the extended Tensor-Based Neural Network (TBNN) employed in \cite{tian_physics-informed_2021} which builds upon the TBNN framework introduced in \cite{ling_reynolds_2016}, with enhancements to incorporate skew-symmetric tensor bases.}
	\label{fig:tbnn}
\end{figure}

\cite{tian_physics-informed_2021} suggested to represent ${\bm H}$ as a function of ${\bm m}$ via a  {  DNNs} trained on the data from the Direct Numerical Simulations (DNS) of turbulence. This is why we call the approach -- {\bf Lagrangian Neural Closure} (LNC) for the pressure Hessian. Moreover, and utilizing earlier ideas of \cite{pope_more_1975}, also developed by \cite{ling_reynolds_2016}), the DNN-parameterized function (see Fig.~(\ref{fig:tbnn})) was constrained to obey rotational symmetry of the homogeneous isotropic turbulence symmetries, then resulting, according to the Cayley-Hamilton theorem, in the finite representation via the tensor basis:
\begin{align}\label{eq:h-NN}
& {\bm h}=\sum_{n=1}^{10} g^{(n)}(\lambda_1,...,\lambda_5){\bm T}^{(n)},\ 
\lambda_1=\text{tr}[{\bm s}^2],\ \lambda_2 = \text{tr}[{\bm w}^2],\ \lambda_3 = \text{tr}[{\bm s}^3],\
\lambda_4=\text{tr}[{\bm w}^2{\bm s}],\  \lambda_5 = \text{tr}[{\bm w}^2{\bm s}^2],\\ \nonumber 
& {\bm T}^{(1)}={\bm s},\ {\bm T}^{(2)} = {\bm s}{\bm w} - {\bm w}{\bm s},\ 
{\bm T}^{(3)} ={\bm s}^2-\frac{1}{3}{\bm I}\text{tr}[{\bm s}^2],\ {\bm T}^{(4)} ={\bm w}^2-\frac{1}{3}{\bm I}\text{tr}[{\bm w}^2],\
{\bm T}^{(5)} =({\bm w}{\bm s})^2-{\bm s}^2{\bm w},\\ \nonumber & {\bm T}^{(6)} ={\bm w}^2{\bm s}+
({\bm s}{\bm w})^2-\frac{2}{3}\text{tr}[({\bm s}{\bm w})^2],\
{\bm T}^{(7)} =({\bm w}{\bm s}{\bm w})^2-{\bm w}^2 {\bm s}{\bm w},\  {\bm T}^{(8)} =
({\bm s}{\bm w}{\bm s})^2-{\bm s}^2 {\bm w}{\bm s},\\ \nonumber &
{\bm T}^{(9)} ={\bm w}^2{\bm s}^2+{\bm s}^2 {\bm w}^2-\frac{2}{3}{\bm I}\text{tr}[{\bm s}^2{\bm w}^2],\ {\bm T}^{(10)} =({\bm w}{\bm s})^2{\bm w}^2 - {\bm w}^2{\bm s}^2{\bm w}, 
\end{align}
where the $(3\times 3)$ matrix/tensor ${\bm m}$ is split in symmetric and skew-symmetric components, ${\bm m}={\bm s}+{\bm w}$ -- strain matrix ${\bm s}$ and vorticity matrix ${\bm w}$, respectively; 
$\lambda_{1,\cdots,5}$ are five scalar invariants; and ${\bm T}^{(1,\cdots,10)}$ are ten tensor bases. Similar DNN approximations/closures were developed for the injection and dissipation terms in Eq.~(\ref{eq:VG}). 

The capability of this method (from \cite{tian_physics-informed_2021}) to extrapolate  to Reynolds numbers larger than those used in training has recently been demonstrated in \cite{buaria_forecasting_2023}.

In summary, \cite{tian_physics-informed_2021}) anchored the deterministic segment of the phenomenological  model in the local approximation closures while incorporating core physical constraints -- Galilean invariance (enforced automatically in the Lgrangian frame), rotational invariance, and the incompressibility condition. These principles were explicitly integrated into Eq.~(\ref{eq:VG}), with modifications such as replacing ${\bm h}$ according to Eq.~(\ref{eq:h-NN}), and analogous alterations for injection and dissipation terms. The model was further enhanced by a stochastic term with zero mean, representing the impact of scale fluctuations on the principal viscous scale interactions. The approach harnessed {  DNNs} to capture functional degrees of freedom, like $g^{(n)}(\cdot)$ from Eq.~(\ref{eq:h-NN}), through training on Lagrangian data derived from high-Reynolds number DNS. The validation process involved a rigorous out-of-sample test, which demonstrated the model’s enhanced ability to represent the magnitude and orientation of the small-scale pressure Hessian contributions. The joint probability density function of the second and third invariants of the Velocity Gradient Tensor was in a satisfactory concordance with the DNS 'ground-truth' data. Moreover, the model successfully captured several key turbulence structures, even those not explicitly embedded within it.

However, it is also important to emphasize that juxtaposition against the ground truth data has also identified challenges, presumably due to limitations of the LNC, in modeling inertial range dynamics, which indicates that a richer modeling strategy is required. (We further elaborate on directions for future research, in particular towards including inertial range geometry into the Tensor Basis DNN approach, in the discussion closing this Section of the manuscript.) 

\subsection{Evolution of AI for Turbulence (Eulerian and also Lagrangian)} \label{sec:Eulerian}

In the forthcoming subsections, we will further explore the use of AI to develop a Lagrangian perspective on turbulence and to effectively utilize Lagrangian data. Subsequently, our attention will pivot to multi-particle (Lagrangian) Neural Closures, which represent a departure from the four-particle Neural Closure examined so far.

However, before delving deeper into these "AI for Lagrangian Turbulence" approaches, it is worthwhile to briefly and, with a certain degree of author bias, review other AI-based methods that have significantly advanced our understanding of turbulence over the past decade. Specifically, in this subsection, we will provide a concise overview and attempt to categorize the methods that mainly adopt an Eulerian perspective of turbulence.

We continue our discussion of {\bf Neural Closure} approaches, akin to those previously explored in the context of Lagrangian turbulence for the pressure Hessian, but now applied to features stemming from classic approach in computational fluid mechanics related to Eulerian closures. Initially, these papers -- which we keep under the rubric of the Eulerian Neural Closures (ENCs) -- aimed primarily at uncovering dependencies necessary for deriving simplified engineering models of turbulence, such as Reynolds Averaged Navier-Stokes (RANS) models \cite{Tennekes1978AFirst} and Large Eddy Simulation (LES) models \cite{sagaut_large_2001}, as reviewed in \cite{duraisamy_turbulence_2019}. An illustrative example of this approach is found in \cite{wang_physics-informed_2017}, where the "physics" was limited to modeling Reynolds stresses -- a crucial aspect necessitating closure within the RANS models -- as a function of the mean flow (and related characteristics), which were then parameterized using a DNN. Similar methodologies were subsequently developed in a series of papers beginning with \cite{wang_investigations_2018}, as well as in \cite{park_toward_2021} and related references, to represent the sub-grid stress tensor contributions in the LES models as a function of the resolved features, including velocities (and occasionally velocity gradients and higher derivatives) at various points on the computational (Eulerian) grid of the LES. ENC approaches were also extended to explore inclusion of physics through representations. For instance, \cite{ling_reynolds_2016} proposed using DNNs to learn coefficients in the Tensor Basis expansion for the Reynolds Stress anisotropy tensor, $\mathbb{E}[v'_i v'_j] - \delta_{ij}$, where $\bm{v}' = \bm{v} - \mathbb{E}[\bm{v}]$.

The application of physically grounded representations has expanded beyond the realm traditionally associated with Neural Closures. For instance, \cite{mohan_wavelet-powered_2019} proposed using Long Short-Term Memory (LSTM) networks, which can be regarded as DNN-enhanced regression models from classical machine learning, to capture the spatio-temporal evolution of turbulent fields (such as the velocity field) through wavelet decomposition.

The employment of LSTM in turbulence modeling signifies a noteworthy transition in the evolution of AI models for turbulence—from merely using DNNs for function approximation to the adoption of generative models. Both regression models and LSTMs are now considered precursors to contemporary generative AI models, illustrating a broader trend towards more sophisticated, generative approaches in the field.

The incorporation of Generative Adversarial Networks (GANs) in turbulence research marks the onset of a new era in generative AI. \cite{goodfellow_generative_2014} pioneered this approach, which was extended by \cite{king_deep_2018} to generate snapshots of the velocity field. This trajectory of research, encompassing the use of LSTM, GANs, and Diffusion Models, can be collectively referred to as {\bf Generative Turbulence} (GenTurb).

In the latest advancements of the Generative Turbulence (GenTurb) methodology, diffusion models have been employed to supplement, enhance, or generate samples of turbulent flows in both Eulerian and Lagrangian frameworks, as detailed in \cite{li_dissecting_2023} and \cite{li_synthetic_2023} respectively.

Methodological contribution of \cite{king_deep_2018} consisted in a suggestion that the physics of turbulence could be retrospectively analyzed by examining the symmetries, constraints, and relationships inherent in the ground truth data, which are not explicitly integrated into the DNN-based framework. In other words, the methodology calls for a physics-of-turbulence-informed post-training diagnostic. This diagnostic tool was then employed in \cite{mohan_wavelet-powered_2019,mohan_spatio-temporal_2020}  to evaluate the quality of LSTM models that generate the turbulent velocity field as a time sequence. Subsequently, in \cite{mohan_compressed_2019}, the diagnostic approach was applied to its compressed variant, where the LSTM, interposed between an encoder and decoder, operates within the compressed (latent) space. 

In contrast, the next wave in the AI for turbulence research emphasizes the incorporation of physical symmetries, constraints, and relations, though not directly applying physics-based equations, as a defining feature. This approach is exemplified in \cite{mohan_embedding_2023}, which demonstrates how to enhance the encoder -- LSTM -- decoder architecture from \cite{mohan_compressed_2019} with an additional pre-output layer that explicitly enforces physics constraints, such as incompressibility.

The integration of DNNs directly into the physical equations characteristic of turbulence phenomenology represents another important advancement in the field. \cite{portwood_turbulence_2019} proposed using a Neural Ordinary Differential Equation (Neural ODE) \cite{chen_neural_2019} to model the dynamics of energy dissipation in decaying turbulence regimes. This AI framework involves encoding the data into a latent space, applying a Neural ODE in a general form, and then decoding it back to the physical space, which encompasses physical characteristics like energy dissipation and effective Reynolds number. This approach was seen as a generalization of the $k-\epsilon$ model \cite{launder_numerical_1974}  -- a classical mode in computational fluid mechanics, which can be considered a variant of the RANS model.  

In the forthcoming two subsections, we will delve into more advanced examples of {\bf Neural Turb-ODE}s, where the right-hand side of the ODEs exhibits significant structural complexity, deriving from multi-particle turbulence closures.

This subsection concludes with two examples that diverge from the previously discussed categories of \textbf{Neural Closure}, \textbf{GenTurb}, and \textbf{Neural Turb-ODE}, yet exhibit some similarities. \cite{fonda_deep_2019} suggest using DNNs to compress flow snapshots indicative of convection, utilizing a U-shaped DNN architecture with contraction and expansion branches to distill the complex 3D turbulent structure into a simpler temporal planar representation at the convection layer's midplane. 
{  The U-net architecture was also utilized by \cite{cremades_identifying_2024} to predict the emergence of coherent structures in boundary layer flows.}
\cite{stulov_neural_2021} develop a Neural Particle Image Velocimetry (PIV) technique to reconstruct the velocity field from consecutive snapshots of the density (passive scalar) field. While acknowledging the field's breadth and the emergence of recent works like, 
\cite{han_yuhuan_attention-mechanism_2023} and \cite{yu_robust_2023} which apply transformers to similar ends, we note a commonality between the problems of snapshot compression and velocity reconstruction from scalar field snapshots. Both entail learning an instantaneous mapping between two related but distinct snapshots of physical fields. This shared characteristic suggests grouping these instances (and potentially others) into a category termed \textbf{Neural Maps}.

\subsection{From Smooth Particle Hydrodynamics to Neural Lagrangian Large Eddy Simulation}
\label{sec:multi}

This subsection is devoted to what can be considered as a Lagrangian multi-particle model from the  class of the  {\bf Neural Turb-ODE} model which we started to discuss in the previous subsection. This approach originates from Smooth Particle Hydrodynamics (SPH). The SPH, as introduced and developed in \cite{monaghan1977,monaghan1992,monaghan12}, serves as the Lagrangian counterpart to predominantly Eulerian multi-scale model reduction techniques, such as Large Eddy Simulations (LES), Proper Orthogonal Decomposition \cite{lumley_book_2012}, and Dynamic Mode Decomposition \cite{schmid_2010, dmd_book}. Its application extends across both weakly and strongly compressible turbulence within astrophysics and various engineering fields \cite{shadloo_2015_industry}.

This subsection focuses on Lagrangian multi-particle models within the Neural Turb-ODE framework, extending the discussion from the previous subsection. The genesis/origin of this approach is in Smooth Particle Hydrodynamics (SPH), which is delineated in \cite{monaghan1977,monaghan1992,monaghan12} as the Lagrangian analog to predominantly Eulerian multi-scale model reduction techniques. These include Large Eddy Simulations (LES) \cite{sagaut_large_2001}, Proper Orthogonal Decomposition \cite{lumley_book_2012}, and Dynamic Mode Decomposition \cite{schmid_2010, dmd_book}. SPH has been applied to both weakly and strongly compressible turbulence scenarios in astrophysics and a range of engineering disciplines \cite{shadloo_2015_industry}.

\begin{figure}[h!]
        \centering
        \begin{subfigure}[b]{0.2\textwidth}
        \centering \includegraphics[height=0.8\textwidth]{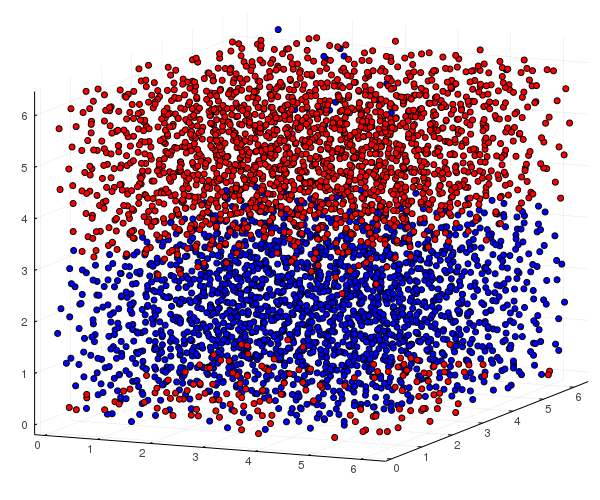}
        \caption{$t_0$}
        \end{subfigure}
        \begin{subfigure}[b]{0.2\textwidth}
        \centering \includegraphics[height=0.8\textwidth]{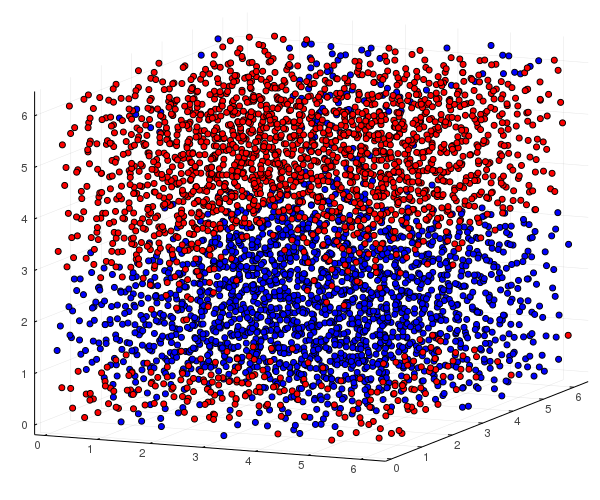}
        \caption{$t_1$}
        \end{subfigure}
        \begin{subfigure}[b]{0.2\textwidth}
        \centering \includegraphics[height=0.8\textwidth]{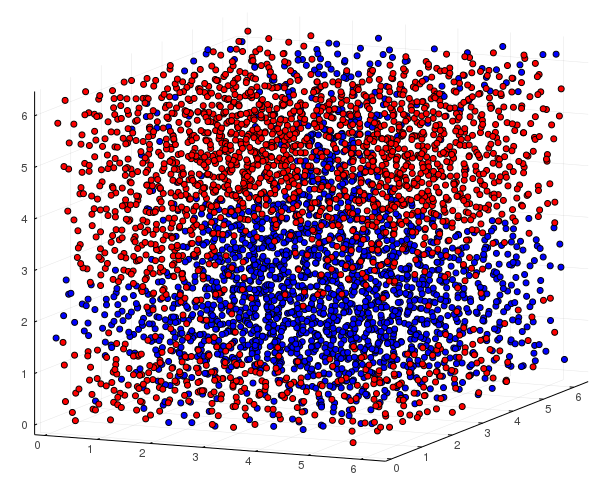}
        \caption{$t_2$}
        \end{subfigure}
        \begin{subfigure}[b]{0.2\textwidth}
        \centering \includegraphics[height=0.8\textwidth]{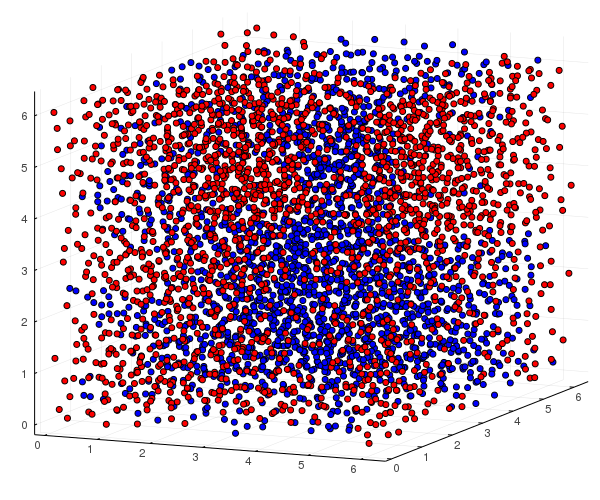}
        \caption{$t_3$}
        \end{subfigure}
        \caption{Temporal evolution, $t_0<t_1<t_2<t_3$, of SPH particles under external forcing $f_{ext}$, utilized as training data to validate the learning algorithm presented in \cite{woodward_physics-informed_2023}. Colors are applied to enhance visualization.}
        \label{fig:sph_flow}
\end{figure}

\cite{tian_lagrangian_2023} and \cite{woodward_physics-informed_2023}, which are collectively reviewed here, relied on SPH as a foundation for AI enhancements, attributing the choice to the SPH distinctive and attractive Lagrangian (multiple-particle) nature among other reduced turbulence models. This nature is crucial for revealing correlations at the resolved scale, obscured by larger scale eddies \cite{1964Kraichnan,1965Kraichnan}, and for its inherent capability to conserve mass, energy, and momentum irrespective of the resolution.

Essentially, SPH approximates the Navier-Stokes Eqs.~(\ref{eq:NS}) by a set of ODEs for a series of discrete particles, serving as interpolation points, as illustrated in Fig.~(\ref{fig:sph_flow}). For each particle $i$ among $N$ particles, the equations are as follows:
\begin{gather}\label{eq:SPH}
    \frac{d\bm{r}_i}{dt} = \bm{v}_i,\quad  
    \frac{d\bm{v}_i}{dt} = -\sum _{j \neq i}^N m_j \left( \frac{P_j}{\rho_j^2} + \frac{P_i}{\rho_i^2} + \Pi_{ij} \right) \nabla _i  W_{ij}  + \bm f_{ext},
\end{gather}
where $W_{ij} = W(||\bm r_i - \bm r_j||, h)$ is the smoothing kernel, with $P_i$ and $\rho_i$ denoting the pressure and density at particle $i$, respectively. $\Pi_{ij}$ represents the artificial viscosity term, and $\bm f_{ext}$ is the external large-scale forcing. The density and pressure (for the weakly compressible formulation) are computed by $\rho_i = \sum_j m_j W(|\bm r_i - \bm r_j|, h)$, and the ideal gas equation of state $P(\rho) = \frac{c^2 \rho_0}{\gamma} \left[ \left(\frac{\rho}{\rho_0} \right)^{\gamma} - 1 \right]$, respectively.

The central concept in \cite{tian_lagrangian_2023,woodward_physics-informed_2023} is to generalize the structure of the right-hand sides (rhs) of the SPH Eqs.~(\ref{eq:SPH}), thereby granting increased flexibility in the physical and interpretable parameters while also employing DNNs to represent functional degrees of freedom.

\cite{woodward_physics-informed_2023} proposed a spectrum of models -- which we call {\bf Neural Lagrangian} models -- extending from a nearly pure Neural ODE, where the rhs of the SPH equations is substituted by a DNN (also with some symmetries accounted for), to a generalized SPH framework. In this framework, the kernel $W(\cdot)$, equation of state $P(\cdot)$, and the eddy-viscosity $\Pi(\cdot)$ are selected in predefined forms, parameterized by readily interpretable parameters.

\cite{tian_lagrangian_2023} focused on developing a model, termed the {\bf Lagrangian LES} (L-LES) model, which allows for a flexible compromise between the structured SPH framework and the unstructured Neural ODE approach. This balance is achieved while maintaining consistency with physical symmetries such as conservation of mass and momentum, and local homogeneity and isotropy. The L-LES generalizes the SPH Eqs.~(\ref{eq:SPH}) as follows:
\begin{eqnarray}
\frac{d\bm{\phi}_{i}}{dt}=\frac{d}{dt} \begin{bmatrix} \rho_i \\ \bm{v}_i \end{bmatrix} &=&
\sum_{j=1}^{N} \begin{bmatrix} \mathcal{N}_\rho (I_{ij,m}, m \in 1,...,5; \lambda_\rho )\\ \displaystyle \sum_{k=1}^{2} \mathcal{N}_{v,k}(I_{ij,m}, m \in 1,...,5; \lambda_v ) \bm{b}_{ij,k}+\Pi_{ij}\bm{b}_{ij,1} \end{bmatrix} + \begin{bmatrix} 0 \\F_i \end{bmatrix},
\end{eqnarray}
where $\mathcal{N}_\cdot(\cdot)$ represents parameterized DNNs. $I$ and $\bm{b}$ are characteristics, normalized to their respective root-mean-square (rms) values, that depend on the positions, velocities, and local densities associated with particle pairs:
\begin{eqnarray*}
\bm{x}_{ij} &=& (\bm{x}_i - \bm{x}_j)/d, \, \bm{v}_{ij} = (\bm{v}_{i} - \bm{v}_j)/v_{rms}, \, \rho_{ij} = \frac{1}{2} (\rho_i +\rho_j)/\rho_{rms}, \\
I_{ij,1} &=& \rho_i/\rho_{rms}, \, I_{ij,2} = \rho_j/\rho_{rms}, \, I_{ij,3} = \vert \bm{x}_{ij} \vert, \, I_{ij,4} = \vert \bm{v}_{ij} \vert, \, I_{ij,5} = \bm{x}_{ij}\cdot\bm{v}_{ij}, \\
\bm{b}_{ij,1} &=& \bm{x}_{ij}, \, \bm{b}_{ij,2} = \bm{v}_{ij}.
\end{eqnarray*}

The nomenclature {\bf Lagrangian Large Eddy Simulation} (L-LES) for the model warrants clarification. This designation is an analogy to the conventional Large Eddy Simulation (LES), which is framed in terms of the velocity field on an Eulerian grid. In L-LES, however, the approach is reconceptualized to express LES heuristics through the velocities, densities, and pressures of Lagrangian particles that are advected by the flow. In this context, the mean distance between neighboring particles in L-LES analogously functions as the resolved scale in standard LES.

The model presented in \cite{tian_lagrangian_2023} was trained using Ground Truth (GT) data derived from {  DNS} on a $1,024^3$ Eulerian grid, alongside Lagrangian trajectories of $64^3$ non-inertial fluid particles carried by the DNS flow. The dataset employed in \cite{woodward_physics-informed_2023}, while smaller in scale, was fundamentally similar in its approach and structure.

\begin{figure}
    \centering
    \includegraphics[width=5in]{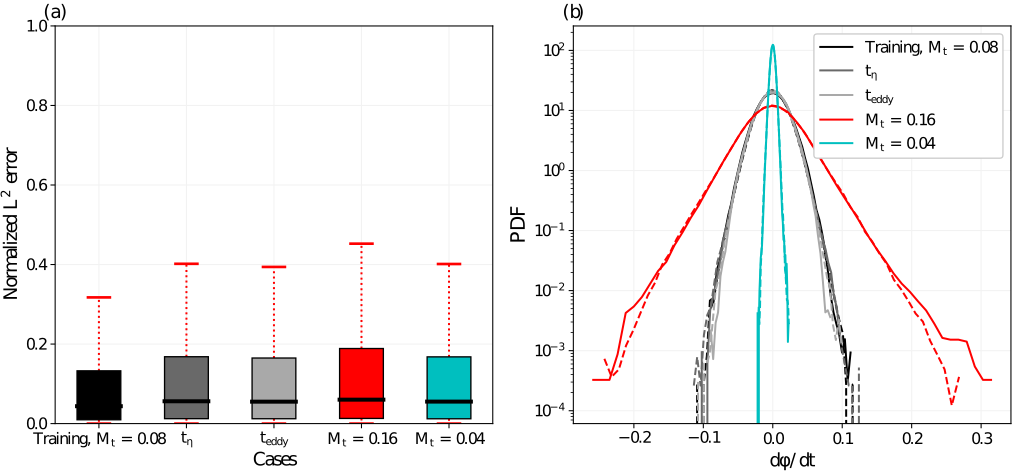}
    \caption{(From \cite{tian_lagrangian_2023}.) Performance on the trained L-LES model in interpolation and extrapolation errors over delayed intervals and different level of turbulent Mach number, respectively. (a) shows normalized $L_2$ error; and (b) shows Probability Distribution Functions (PDF) of the prediction.}
    \label{fig:error_generalization}
\end{figure}

In \cite{tian_lagrangian_2023}, the training occurred in two phases: initially, a smoothing kernel, ${\bm W}(\cdot)$ represented by a DNN was trained, followed by the training of DNNs for the rhs of the equations. The loss function for the smoothing kernel combined terms to ensure alignment with the GT data for velocity and its gradient (coarsely integrated over the inter-particle distance on the Eulerian grid). The DNNs' training employed a loss function comprising three distinct parts, representing (a) time-integrated matches of densities, velocities, and velocity gradients at Lagrangian points; (b) analogous matches at Eulerian grid points; and (c) Kullback-Leibler (KL) matching of instantaneous single-particle velocity and gradient distributions. This elaborate loss function and two-stage training process aimed to stabilize the computational scheme and permit validation-stage experiments on result sensitivity to weighting of different terms in the loss function.

\begin{figure}[h!]
\centering
\begin{subfigure}[b]{0.8\textwidth}
\centering
\includegraphics[width=0.9\textwidth]{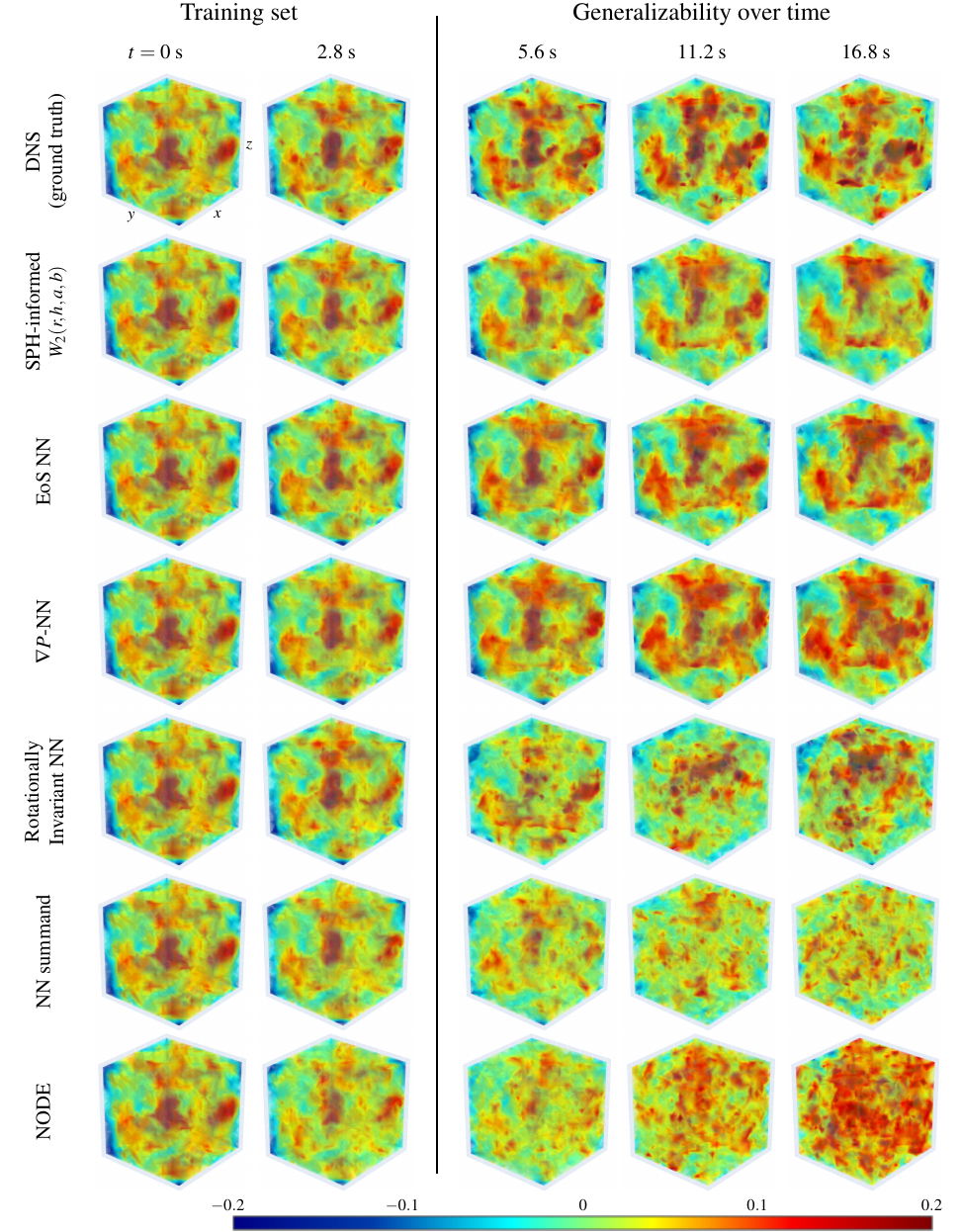}
\end{subfigure}
\caption{Volume plots comparing the Eulerian $u$ velocity component from coarse-grained DNS data with model predictions from \cite{woodward_physics-informed_2023}. These illustrate improved model generalization to the longest physically relevant time scale, $t_{eddy}$, especially with increased SPH structure inclusion, despite training on the shortest scale $t_{\eta}$. The large-scale structures in the $u$ velocity are better captured by SPH-informed models, with performance diminishing as DNN-based acceleration parameterization increases.}
\label{fig:gen_t_qual_dns}
\end{figure}

In contrast, \cite{woodward_physics-informed_2023} {  trained the} {\bf Neural Lagrangian} models in a single stage using a loss function that integrated two main components: (a) matching the velocity field across the Eulerian grid and (b) KL matching of single-particle Lagrangian velocity distributions. Additionally, \cite{woodward_physics-informed_2023} developed a mixed-mode method combining local sensitivity analysis with forward and backward AD to efficiently compute gradients of the loss function across both physical and DNN-embedded parameters. \footnote{  AD can be performed in two primary modes: forward and reverse. Forward mode AD is efficient for functions with fewer inputs and more outputs, propagating derivatives alongside the original function evaluation. In contrast, reverse mode AD, also known as backpropagation, is optimal for functions with many inputs and a single output, making it particularly useful for DNNs. In the context of the Neural Lagrangian models, reverse mode AD facilitates the optimization of the loss function by efficiently computing gradients with respect to both physical and DNN-embedded parameters. For further insight on modern ways of utilizing AD in learning-based, simulation-driven scientific discovery, refer to \cite{lavin_simulation_2022}.}

The insights and subsequent applicability of \cite{woodward_physics-informed_2023} and \cite{tian_lagrangian_2023} are mutually reinforcing. The range of Neural Lagrangian models presented in \cite{woodward_physics-informed_2023} holds significant relevance for practitioners, such as engineers and astrophysicists, who depend on SPH modeling. This work provides a roadmap for progressively refining SPH models to align more closely with Ground Truth (GT) data. See Fig.~(\ref{fig:gen_t_qual_dns}) for illustration. Conversely,  and as illustrated in Fig.~Fig.~(\ref{fig:error_generalization}), \cite{tian_lagrangian_2023} focused on showing how the L-LES model facilitates extrapolation into domains not represented in the GT data. Notably, it was demonstrated that, despite training on relatively brief episodes corresponding to the turbulence turnover time at the resolved scale, L-LES can extrapolate over longer periods, yielding statistically robust predictions. This capability is evidenced through post-factum tests, which include statistics of the invariants of the velocity gradient and the statistics of Lagrangian acceleration, both coarse-grained at the resolved scale.

Additionally, several approaches have emerged that fit into the {\bf Neural Lagrangian modeling} category. Aligning with Neural ODE end of the Neural Lagrangian modeling  spectrum as outlined in \cite{woodward_physics-informed_2023}, \cite{ummenhofer_lagrangian_2020} introduced a framework that interprets the right-hand side of equations for Lagrangian particle positions through continuous spatial convolutions. This concept was further developed in \cite{prantl_guaranteed_2022}, demonstrating how to adapt the Neural Lagrangian modeling framework to ensure momentum conservation for each particle. More recently, \cite{winchenbach_symmetric_2024} expanded upon the continuous convolution concept from \cite{ummenhofer_lagrangian_2020} by integrating separable basis functions, thereby creating a comprehensive method that encompasses traditional techniques such as Fourier-based approaches.

\subsection{AI for Turbulence: What is Next?} \label{sec:turb-next}

In previous segments, we've explored a diverse array of AI-driven methods that are suggesting new AI venues for  turbulence research. This concluding subsection complements the material and casts my vision forward --- strategic and tactical -- spotlighting emerging concepts and platforms that I find promising.

Beginning with the long-term, I aim to extend the AI perspectives beyond merely comprehending turbulence from a theoretical perspective. I would like to seek to master  controlling turbulence in practical, engineering applications, such as advancing fusion technology and enhancing aircraft and rocket designs and/or understanding and improving control of complex multi-physics flows, such as related to stochastic transitions from laminar to turbulent in polymer solutions. On a more tactical level, particularly within the realm of predominantly Lagrangian methods, my long-term objectives are likely to be realized through strategies that emerge as subsets or hybrids of the following approaches:

\begin{itemize}
\item {\bf Reduced Models – New Interactions and Novel Interpretations:} We have explored the expansion of the SPH framework, proposing models that bridge the gap from over-determined SPH to underdetermined DNN-based models, with fundamental symmetry constraints. This approach opens avenues for various hierarchical structures. For instance, we could evolve towards complex equations that include summations over higher-order interactions, such as triplets of particles, rather than just pairs. Additionally, the potential to explore nuanced interactions between fields and particles allows for experimentation with transitions between these states and their implications. Such investigations could lead to the generalization of kernels and the establishment of novel hierarchies. Concurrently, reevaluating the loss functions to adjust the emphasis on different model components presents an opportunity. Moreover, reconsidering the training methodology to potentially integrate the entire process into a unified optimization framework moves away from the two-step training approach employed so far in \cite{tian_lagrangian_2023}.

\item {\bf Extrapolation -- Exploring the Unknown:} The aspiration in Scientific AI is to train within accessible data realms and then extend into specialized interest areas, such as high Reynolds number flows or rare events like once-in-a-lifetime hurricanes. The ideal approach to exploring these new regimes should be gradual, bridging the gap between the "standard/interpolated" and "non-standard/extrapolated" in an adiabatic manner, avoiding abrupt transitions. Innovations in AI, particularly with diffusion and transformer models, are anticipated to play a crucial role here. We envisage a transition between explored and unexplored regimes using an artificial, possibly stochastic, bridge-diffusion dynamics. Another promising strategy involves employing reinforcement learning to navigate challenging domains \cite{mojgani_extreme_2023,sorensen_non-intrusive_2024}, where the reward function is tailored to subtly bias toward the rare regimes.

\item {\bf Lagrangian Multi-Fidelity -- Merging Tetrad and Multi-particle Approaches:} In the vein of \cite{pawar_multi-fidelity_2021}, we aim to develop a hierarchy of Physics-Informed models varying in fidelity, encompassing different levels of coarse-graining and modalities. Specifically, we propose to extend the concept to include more complex quasi-particles, such as tetrads, offering richer dynamics than singular particles. Moreover, integrating both Eulerian and Lagrangian modalities within the models -- not just within the ground truth (GT) datasets -- offers a promising direction for comprehensive multi-fidelity modeling.

\item {\bf Less Data -- More Physics:} I suggest that the integration of physics-guided principles with diffusion models can significantly enhance data compression in multi-scale scientific challenges, thereby improving efficiency and accuracy. Leveraging the advancements in diffusion, transformers, and reinforcement learning (RL) discussed in this paper, along with other cutting-edge AI methods, we aim to refine these models for optimal data compression. This enhancement should maintain or even augment the integrity and interpretability of critical features, akin to the Lagrangian LES models previously mentioned. Furthermore, adopting an RL approach to data reduction, by assigning a "price tag" to data based on its utility for specific physical phenomena or tasks in turbulence research, could prioritize data in a way that enhances data acquisition efficiency. Integrating Uncertainty Quantification (UQ) methods should further strengthen performance evaluation and support more strategic decision-making in both the discovery and application aspects of turbulence studies.

\item {\bf Diffusion Models of AI with Physical Time Line:} Aligning with \cite{holzschuh_score_2023}, adapting diffusion models to physical time processes appears highly promising, transitioning from artificial to physical time dynamics. In this context, expanding the Lagrangian perspective on turbulence, as detailed in previous sections, to encompass fluid blob dynamics that transport correlations from large integral scales to smaller scales of interest in the inertial range, is a viable path. {  Notice that most recently an interesting step in this direction was taken in \cite{li_synthetic_2024,li_generative_2024}, applying diffusion models to the Lagrangian evolution of particles in turbulent flows. A machine learning diffusion model that generates single-particle trajectories was developed. Even though this diffusion model was not physics-informed, it seems to reproduce statistical benchmarks of Lagrangian turbulence and even demonstrate surprisingly strong generalizability for extreme events. Further development of this approach and its integration with physics-informed modeling, including Eulerian field statistics like velocity or scalar fields, should further facilitate the integration of the diffusion modeling into turbulence research.}

{ 
\item {\bf Transformer Models in Turbulence Modeling:} Transformers, as we briefly discussed in Section \ref{sec:science-AI}, represent a branch of generative modeling that has led to some of the most remarkable recent advancements in AI, particularly in the realm of Large Language Models (LLM). \cite{zhu_fluid-llm_2024} introduced Fluid-LLM, which tokenizes a snapshot and then a sequence of snapshots of the fluid flow, using it as a query to generate predictions for how the flow evolves. \cite{solera-rico_beta-variational_2023} applied $\beta$-Variational Autoencoders (which can be viewed as early versions of transformers) and transformers themselves for reduced-order modeling of fluid flows, highlighting these models' capability to efficiently capture complex flow dynamics with reduced computational costs. \cite{yousif_transformer-based_2023} developed a transformer-based synthetic-inflow generator specifically designed for spatially developing turbulent boundary layers. Their work demonstrated the effectiveness of transformers in generating high-fidelity inflow conditions, which are crucial for accurate turbulence simulations. Notably, these applications of transformer-related technology are Eulerian by construction. We envision extending transformer technology to the Lagrangian frameworks, such as the Neural Lagrangian Large Eddy Simulation discussed in Section \ref{sec:multi}, by using Lagrangian quasi-particles as tokens.
}

\item  {\bf Focus on Super-Resolution:} Beyond capturing the large-scale structures of turbulence, there is a compelling need to accurately sample the smaller-scale structures, presupposing the larger ones are known. In AI/ML style, this integration within turbulence research is aptly termed as ``super-resolution" of the subgrid turbulence phenomena, as discussed in \cite{swaminathan_ai_2023} and the references therein. Developing a Lagrangian multi-particle approach for super-resolution analysis presents an intriguing and novel challenge in the field.

\item  {\bf Physics-Informed Reinforcement Learning for Swimmers:} In our discussions, Lagrangian particles were considered synthetic (quasi-particles). Envisioning a statistical description of turbulence where the Lagrangian perspective is embodied by actual particles, some or all of these particles could exert forces to swim, aiming to maintain a formation or stay within proximity of a designated particle or a specific part of the flow. Recent advancements in analyzing and reinforcing optimal swimming strategies in flows, especially turbulent ones, using DNNs have been significant \cite{reddy_glider_2018,rabault_deep_2020,alageshan_machine_2020,borra_optimal_2021,garnier_review_2021,gunnarson_learning_2021,kicic_adaptive_2023,novati_automating_2021,borra_reinforcement_2022}. Yet, the RL methodologies employed have generally lacked or minimally incorporated physical insights from traditional Lagrangian approaches. Thus, developing Physics-Informed RL that integrates reduced Lagrangian modeling for physical guidance presents yet another compelling direction for future research. Initial steps in this direction were taken in \cite{koh_physics-informed_2024}.

\item {  {\bf New Hypotheses:} Turbulence is often referred to as the last unsolved problem of classical mechanics. Over the multi-century history of the field, many hypotheses have been proposed. While the majority have been proven wrong, quite a number were abandoned simply because we lacked the necessary validation tools. With new AI tools at our disposal, we should re-examine these hypotheses. Some of our discussions above, notably in Section \ref{sec:turb}, can be viewed as first steps in this direction. Another interesting example of this type involves developing a famous hypothesis by Hopf made in \cite{hopf_mathematical_1948}, suggesting that turbulence can be explained in terms of the unstable simple invariant solutions populating the inertial manifold of a high-dimensional space. This idea was further explored in \cite{page_recurrent_2024}, where it was shown how to reconstruct PDFs of dissipation rate, production rate, and energy in developed two-dimensional turbulence from a set of unstable periodic orbits. A combination of AI techniques -- AD and DNN (of the deep convolutional autoencoder type) -- as well as Markov chain approaches were utilized to find hundreds of unstable periodic orbit solutions and to build the PDFs. We expect to see more approaches in this spirit, using AI to reconsider old and validate new bold hypotheses about turbulence.
}

\end{itemize}

%\bibliography{bib/DiffusionModels.bib,bib/StatMechForAI.bib,bib/links.bib, bib/VGTetrad.bib,bib/PassiveTurb.bib,bib/MELT.bib,bib/SPH-paper.bib,bib/NeuralPIV.bib,bib/EqsFromData.bib,bib/Transformers.bib,bib/MishaPapers.bib,bib/RLFluids.bib,bib/RareEvents.bib}
%\bibliographystyle{tmlr}

\end{document}